\documentclass[10pt,journal,compsoc]{IEEEtran}

\ifCLASSINFOpdf
\else
\fi
\usepackage{cuted}
\usepackage{hyperref}
\usepackage{url}
\usepackage{multirow}
\usepackage{makecell}
\usepackage[utf8]{inputenc} 
\usepackage[T1]{fontenc}    
\usepackage{hyperref}       
\usepackage{url}            
\usepackage{booktabs}       
\usepackage{amsfonts}       
\usepackage{nicefrac}       
\usepackage{microtype}      
\usepackage{xcolor}         
\usepackage{graphicx}
\usepackage{subfigure}
\usepackage{amsmath}
\usepackage{amssymb}
\usepackage{mathtools}
\usepackage{amsthm}
\usepackage{subfigure}
\usepackage{rotating}

\usepackage{threeparttable}
\usepackage{subcaption}
\usepackage{algorithm}
\usepackage{algpseudocode}
\usepackage{amssymb}
\usepackage{bm}
\usepackage[style=numeric,backend=biber]{biblatex}
\begin{document}
%
\title{TimeCNN: Refining Cross-Variable Interaction on Time Point for Time Series Forecasting}
%
%
%
%

\author{Ao~Hu, Dongkai~Wang, Yong~Dai, Shiyi~Qi, Liangjian~Wen, Jun~Wang, Zhi~Chen, Xun~Zhou, Zenglin~Xu, Jiang~Duan
\IEEEcompsocitemizethanks{
\IEEEcompsocthanksitem Ao Hu, Dongkai Wang, Liangjian Wen, Zhi Chen and Jiang Duan are with Engineering Research Center of Intelligent Finance, Ministry of Education, School of Computing and Artificial Intelligence, Southwestern University of Finance and Economics, Chengdu, 611130. \protect Email: huao1105@gmail.com; wdk@swufe.edu.cn; wlj6816@gmail.com; chenzhi@swufe.edu.cn; duanj\_t@swufe.edu.cn.
\IEEEcompsocthanksitem Shiyi Qi and Xun Zhou are with Harbin Institute of Technology, Shenzhen, 518055. \protect Email: syqi981125@163.com; zhouxun2023@hit.edu.cn.
\IEEEcompsocthanksitem Yong Dai and Jun Wang are with HiThink Research, Hithink RoyalFlush Information Network Co., Ltd., Hangzhou, 310023. \protect Email: daiyongya@outlook.com; wangjun7@myhexin.com.
\IEEEcompsocthanksitem Zenglin Xu is with Artificial Intelligence Innovation and Incubation ($AI^3$) Institute of Fudan University, Shanghai, 201203. \protect Email: zenglinxu@fudan.edu.cn.
\IEEEcompsocthanksitem Liangjian Wen is the corresponding authors.
\IEEEcompsocthanksitem This article is in part by the Sichuan Science Foundation Project (Grants No. 2024ZDZX0002 and 2024NSFTD0054).
}
}

\IEEEtitleabstractindextext{%
\begin{abstract}
Time series forecasting is extensively applied across diverse domains. Transformer-based models demonstrate significant potential in modeling cross-time and cross-variable interaction. However, we notice that the cross-variable correlation of multivariate time series demonstrates multifaceted (positive and negative correlations) and dynamic progression over time, which is not well captured by existing Transformer-based models.
To address this issue, we propose a TimeCNN model to refine cross-variable interactions to enhance time series forecasting. Its key innovation is timepoint-independent, where each time point has an independent convolution kernel, allowing each time point to have its independent model to capture relationships among variables. 
This approach effectively handles both positive and negative correlations and adapts to the evolving nature of variable relationships over time. Extensive experiments conducted on 12 real-world datasets demonstrate that TimeCNN consistently outperforms state-of-the-art models. Notably, our model achieves significant reductions in computational requirements (approximately 60.46\%) and parameter count (about 57.50\%), while delivering inference speeds 3 to 4 times faster than the benchmark iTransformer model.

\end{abstract}

\begin{IEEEkeywords}
Time series forecasting, cross-variable correlationship, timepoint-independent.
\end{IEEEkeywords}}

\maketitle

\IEEEdisplaynontitleabstractindextext

%
\IEEEpeerreviewmaketitle

\IEEEraisesectionheading{\section{Introduction}\label{sec:introduction}}

%
%
%
%
\IEEEPARstart{T}{ime} series forecasting is extensively applied across various fields, including traffic ~\cite{He_Zhang_Bai_Yi_Niu_2022, wen}, finance~\cite{finance_apply, FInance_TKDE}, weather~\cite{weather_apply, Weather_TKDE}, and energy consumption~\cite{RePEc:eee:rensus:v:74:y:2017:i:c:p:902-924}, etc. Deep learning~\cite{DLinear,SCINet,MTGNN,itransformer} methods have exhibited exceptional performance in time series forecasting.
Among them, Transformer-based models~\cite{Informer, Autoformer,fedformer, PatchTST, DNN1_TKDE,DNN2_TKDE} demonstrate significant potential in modeling long-term temporal dependency for cross-time interaction.

In addition to cross-time interaction, recent research increasingly underscores the significance of cross-variable interaction to achieve accurate time series forecasting. 
Some prior works~\cite{STGCN,MTGNN} have leveraged Graph Neural Networks (GNNs) to explicitly capture cross-variable dependency. 
Besides, Informer~~\cite{Informer}, Autoformer~~\cite{Autoformer}, and FEDformer~~\cite{fedformer}  only implicitly utilize this dependency. 
These Transformer-based models embed 
multiple variates of the same timestamp as temporal tokens to capture the dependencies among time points. This interaction of time-unaligned tokens from different multivariate introduces unnecessary noise into forecasting processes. 
 Hence, cross-variable dependency is not captured well.
To explicitly capture multivariate correlations, Crossformer~\cite{Crossformer}  introduces two-stage attention on variables and segments of time points to enhance prediction performance. Recently, iTransformer~\cite{itransformer} embeds
  independent time series as tokens to capture multivariate correlations by self-attention, thereby achieving state-of-the-art (SOTA) performance.

\begin{figure*}[tbp]
\centering
\begin{minipage}[t]{0.48\textwidth}
\centering
\includegraphics[width=\textwidth]{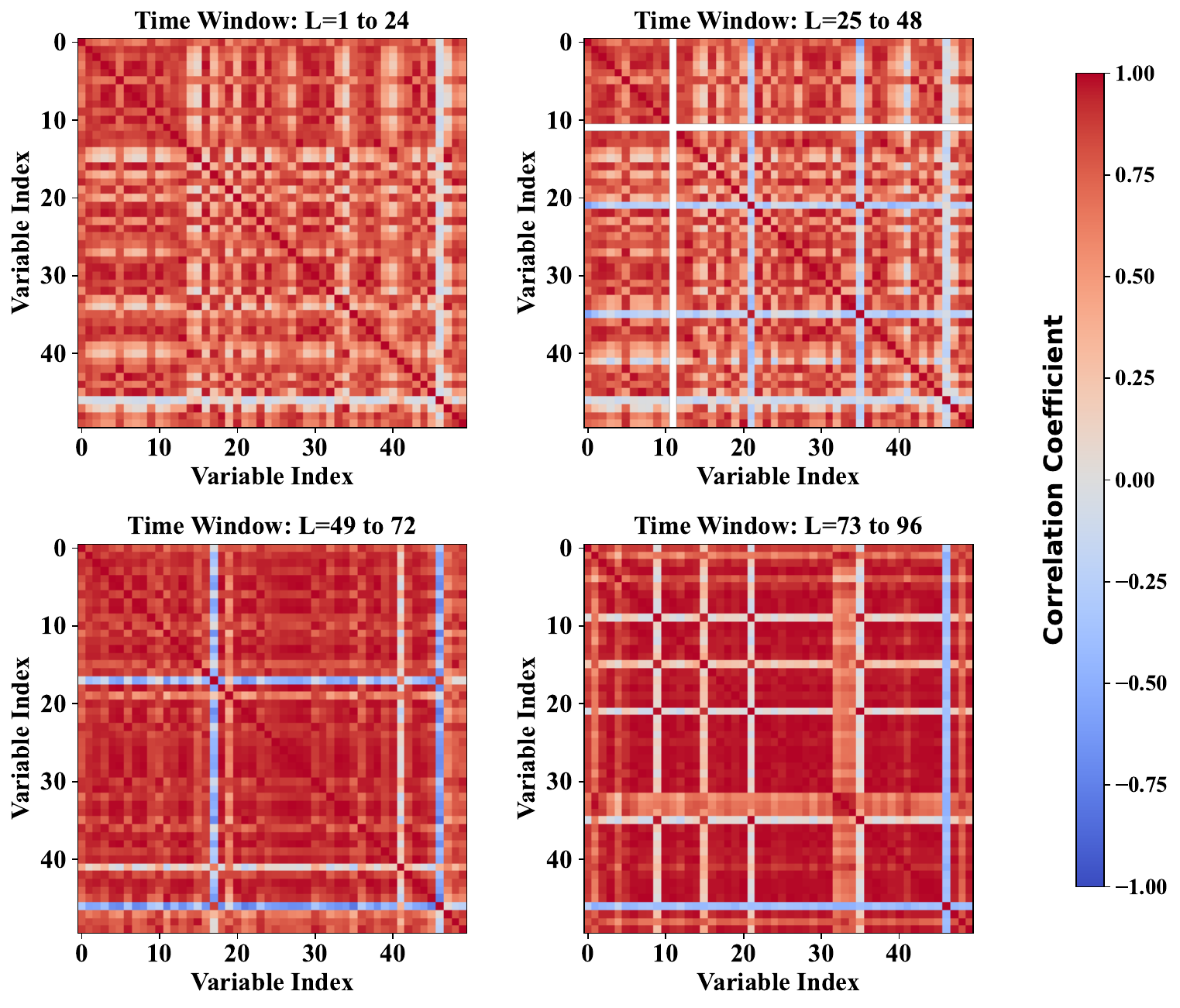}
\caption{A case visualization of variable relationships dynamically changing over time. Pearson correlation coefficients are calculated for each of four equal consecutive segments of the lookback window, using 50 randomly selected variables from the Traffic dataset.}
\label{fig:introduction_traffic}
\end{minipage}\hfill
\begin{minipage}[t]{0.48\textwidth}
\centering
\hspace{-0.3cm}
\raisebox{0.25\height}{\includegraphics[width=\textwidth]{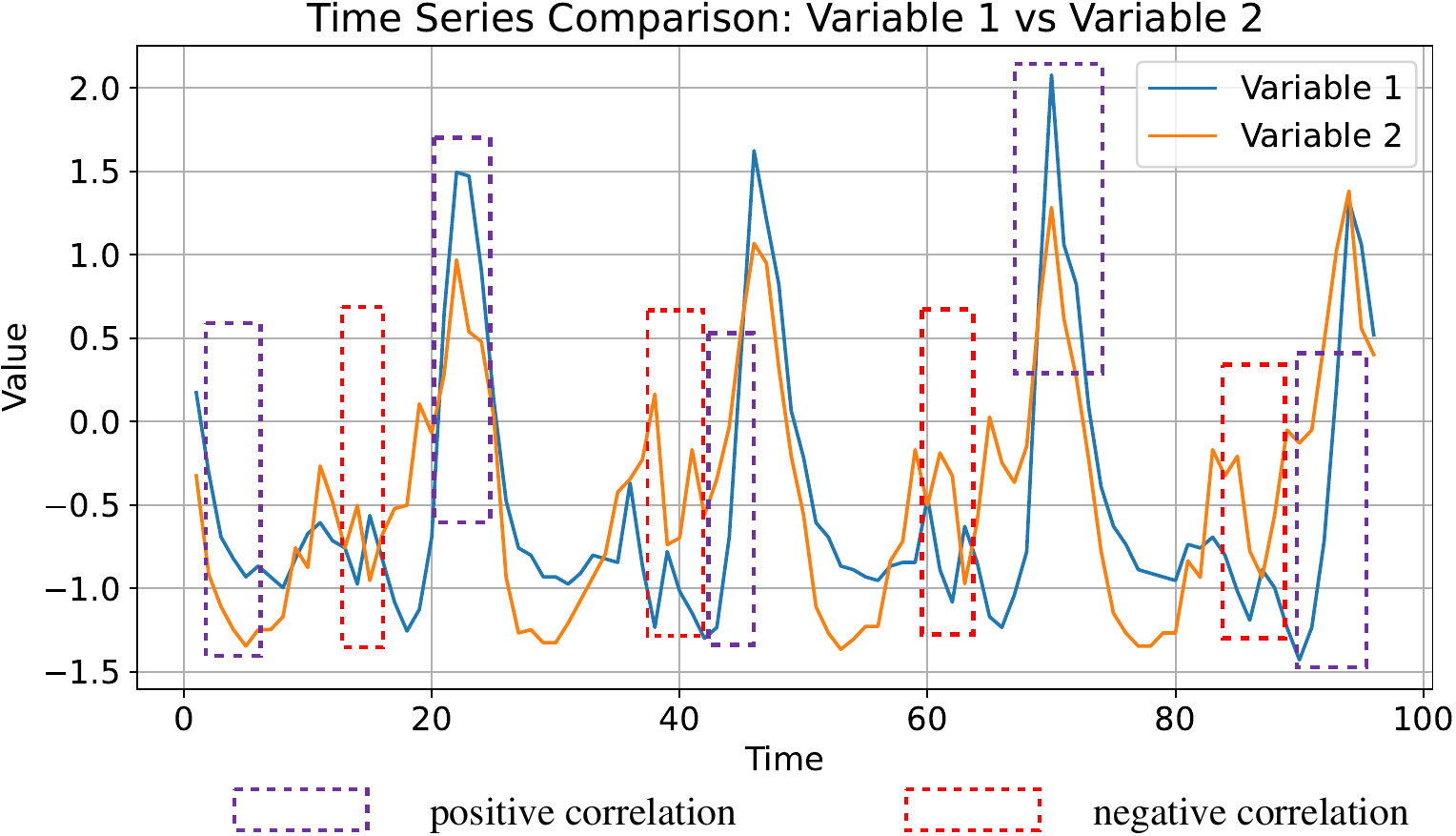}}
\caption{Comparison of two univariable time series from the ECL dataset, with the x-axis representing the lookback window and the y-axis showing variable values at each time step. Purple and red dashed boxes indicate positive and negative correlations.}
\label{fig:variable1_and_variable2}
\end{minipage}
\vspace{-0.5cm}
\end{figure*}



However, iTransformer overlooks the intrinsic dynamic nature of variable relationships over time, as it encodes independent entire time series into variable tokens and captures multivariate correlations among these tokens.  
As shown in Figure~\ref{fig:introduction_traffic}, it is notable that the cross-variable correlation of multivariate time series demonstrates a multifaceted and dynamic progression over time.
This phenomenon
 reflects the complex and evolving nature of the underlying processes and interactions among the variables. Attention among variable tokens in iTransformer cannot portray complex and dynamic multivariate correlation.
 Consequently, this constraint reduces its capacity and generalization ability on diverse time series data. 

\begin{figure}[htbp]
    \centering
    \includegraphics[scale=0.45]{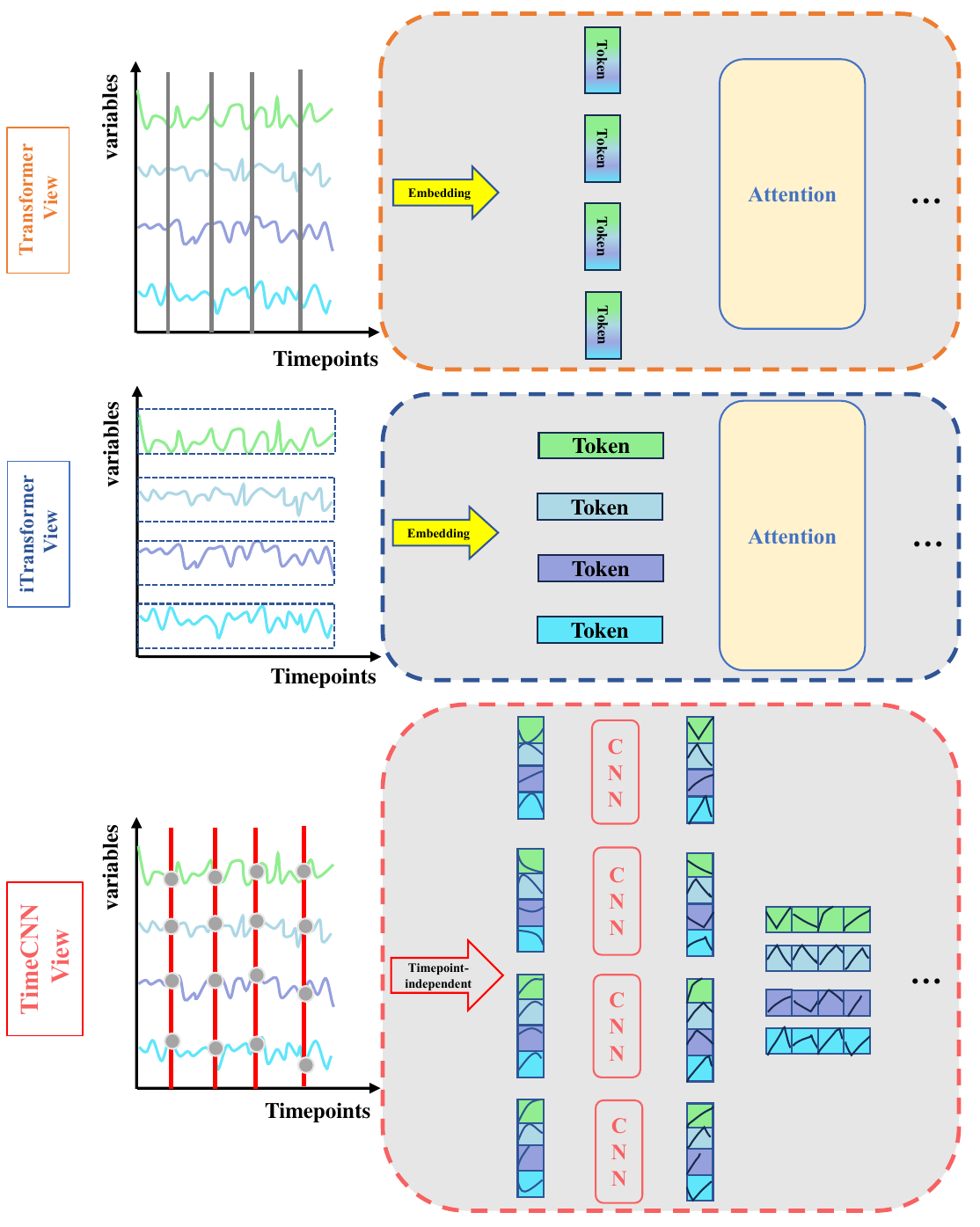}
    \caption{Comparison between the Transformer~\cite{Transformer}, iTransformer~\cite{itransformer}, and TimeCNN. The Transformer embeds all variables at each time point into temporal tokens, using attention to capture cross-time interactions. The iTransformer embeds time points of each variable into variable tokens, leveraging attention for cross-variable interactions. In contrast, our proposed TimeCNN processes each time point independently to capture and model dynamic changes in variable relationships.}
    \label{fig:Intro}
\end{figure}

Additionally, negative correlations are observed in Figure~\ref{fig:introduction_traffic}, indicated by a shift from red to blue. We further illustrate negative correlations in Figure \ref{fig:variable1_and_variable2}.
The purple dashed box indicates that the two variables are positively correlated, while the red dashed box denotes the negative correlation. We notice the variable relationships as they dynamically change over time, exhibiting instances of both positive and negative correlation. More notably, it demonstrates that the relationship of variables between adjacent time points may also change. 
In iTransformer, the learned pre-Softmax maps of the self-attention module also demonstrate similar negative correlations among variables of the entire time series.
However,  negative correlations are transformed into particularly small similarities through the softmax function.
Consequently, the attention mechanism cannot utilize negative correlations to improve time forecasting.

Building on the preceding analysis, refining cross-variable interactions for complex dynamic multivariate correlations remains a challenging task.
To mitigate this issue,  we propose a TimeCNN model that contains the key component: 
\noindent{\textbf{Timepoint-independent.}}    Each time point has an independent convolution kernel. Specifically,  it utilizes a large convolution kernel to capture the relationships within all variables at each time point. 
In this manner, TimeCNN is capable of dynamically capturing the relationships among variables, even when those of adjacent time points differ.
Furthermore, the convolution of all variables at each time point can capture both positive and negative correlations.
A comparison between the iTransformer and TimeCNN is illustrated in Figure~\ref{fig:Intro}. The iTransformer leverages the self-attention mechanism to capture cross-variable interactions among variable tokens. However, it overlooks the dynamic nature of these relationships. In contrast, our TimeCNN captures cross-variable interactions at each individual time point.

\begin{figure}[htbp]
\centering
\includegraphics[scale=0.31]{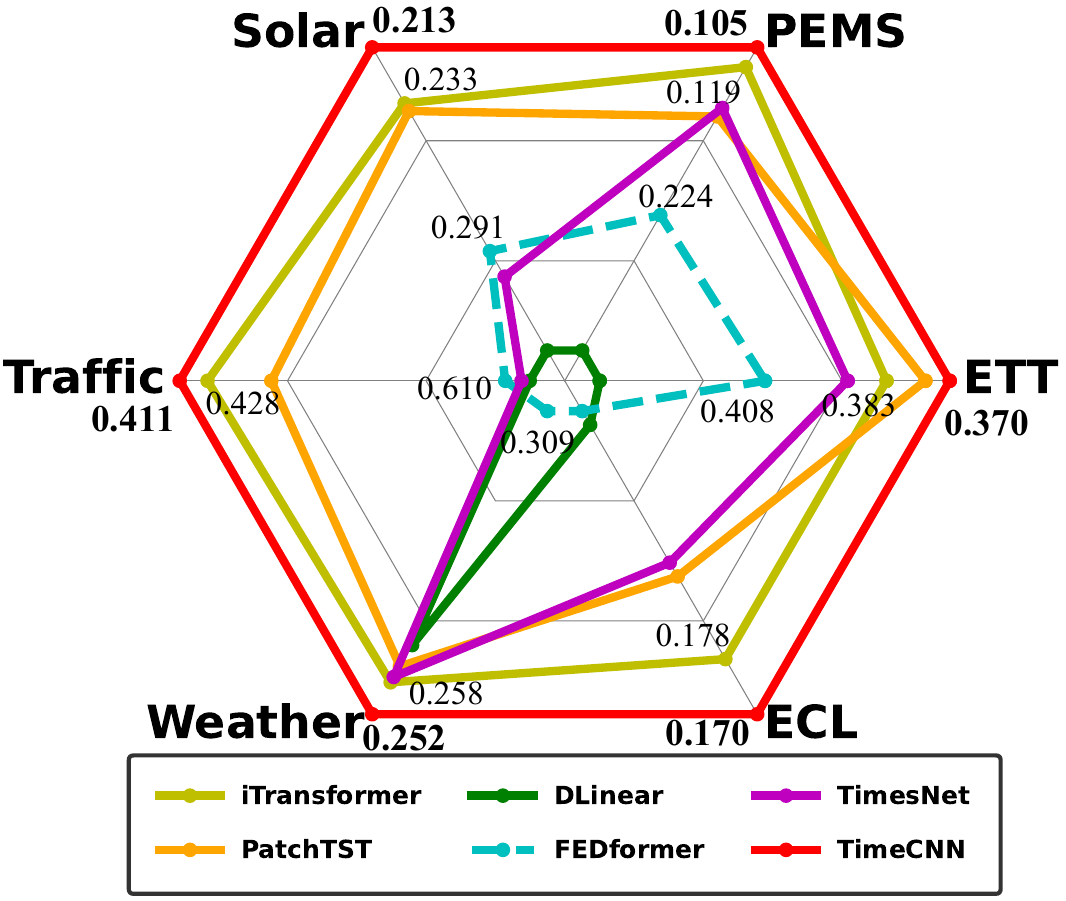}
\caption{Performance of TimeCNN. The results of PatchTST~\cite{PatchTST} and TimeCNN are from our experiments and other average results (MSE) are reported following iTransformer~\cite{itransformer}. }
\label{fig:radar}
\end{figure}

After cross-variable interaction,  TimeCNN embeds the complete time series for each variable and utilizes a feed-forward network to learn time series representation. 
Experimentally, the proposed TimeCNN demonstrates state-of-the-art performance on real-world forecasting benchmarks, as illustrated in Figure \ref{fig:radar}. Remarkably, it tackles the primary challenges associated with refining cross-variable interactions in the context of complex dynamic multivariate correlations. Moreover, compared to iTransformer, TimeCNN achieves a reduction in computational demand by approximately 60.46\% and in parameter conunt by 57.50\%, while enhancing inference speed 3 to 4 times faster. This efficiency can be attributed to the parallel computation of convolutions across all time points.
The main contributions are as follows:
\begin{itemize}
\item 
We observe that Transformer-based models struggle to effectively capture complex dynamic multivariate correlations in time series forecasting. This limits their  generalization capabilities when applied to various time-series datasets.

\item 
We propose TimeCNN to refine cross-variable interactions for complex dynamic multivariate correlations to enhance time series forecasting. Its key component is timepoint-independent, where each time point has an independent convolution kernel.

\item Extensive experimental results on 12 real-world datasets demonstrate that our model outperforms state-of-the-art models. Moreover, compared to iTransformer, TimeCNN achieves a reduction in computational demand by approximately 60.46\% and parameter conunt by 57.50\%, while delivering inference speeds 3 to 4 times faster. 

\end{itemize}
%
%
%
%
\section{Related Work}

Deep learning models have demonstrated remarkable performance in time series forecasting. These models can be broadly divided into Transformer-based models~\cite{Transformer, Autoformer,fedformer, PatchTST, Crossformer, itransformer}, MLP-based models~\cite{DLinear, tide, RLinear}, GNN-based models, and CNN-based models~\cite{SCINet, Timesnet, moderntcn}. These models primarily enhance predictive accuracy by modeling temporal dimensions (cross-time) and variable dimensions (cross-variable).

\begin{figure*}[htbp]
    \centering
    \hspace*{-0.8cm} 
    \includegraphics[scale=0.50]{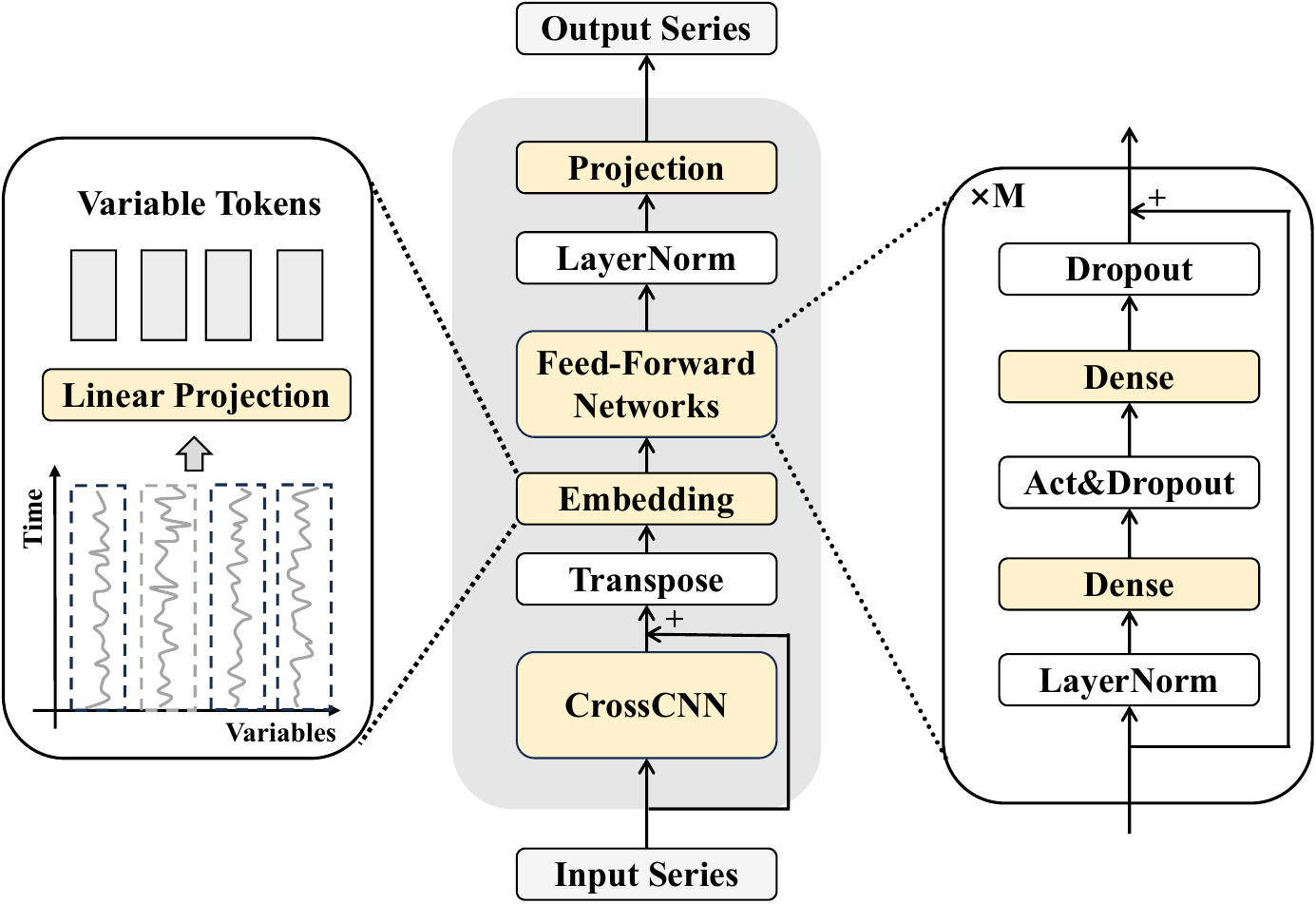}
    \caption{The TimeCNN architecture, which mainly consists of CrossCNN, Embedding, Feed-Forward Networks (FFN), and Projection for prediction. In the CrossCNN module, cross-variable interactions at each time point are captured. Then, the entire time series of each variable is embedded into variable tokens, which are further learned in shared FFN. Finally, the learned multivariate representations are projected through a linear layer to predict future time series.}
    \label{fig:TimeCNN}
\end{figure*}

\subsection{Cross-Time Interaction Models}
Cross-Time Interaction Models primarily capture the dependencies between time points or patches of time points. Among these models, Transformer-based models have shown advancements in time series forecasting owing to their capacity to capture long-range dependencies. Informer~\cite{Informer}, Autoformer~\cite{Autoformer}, and FEDformer~\cite{fedformer} aim to capture temporal dependencies and reduce the quadratic computational cost brought by attention mechanisms. However, their effectiveness is challenged by a simple linear model ~\cite{DLinear}. Additionally, RLinear~\cite{RLinear} verifies that employing the Multilayer Perceptron (MLP) along the temporal dimension can effectively capture periodic features in time series. Subsequently, PatchTST~\cite{PatchTST} relies on a channel-independent strategy to capture dependencies between patches, demonstrating superior performance. In addition, CNN-based models~\cite{BaiTCN2018} capture the information of temporal dynamics by convolutional kernels sliding along the time dimension but face challenges in capturing global dependencies due to the limited receptive field of the convolutional kernels. To overcome this issue, TimesNet~\cite{Timesnet} transforms one-dimensional time series into a set of two-dimensional tensors based on multiple periods and applies two-dimensional convolution kernel modeling to extract complex temporal variations effectively.

\subsection{Cross-Variable Interaction Models}
The increasing emphasis on cross-variable interaction modeling in recent research aims to improve the accuracy of time series forecasting. Previous studies have extensively leveraged Graph Neural Networks (GNNs) to capture cross-variable dependencies. For instance, STGCN~\cite{STGCN} effectively models multi-scale traffic networks, thereby capturing both spatial and temporal correlations. Similarly, MTGNN~\cite{MTGNN} employs GNNs to extract uni-directed relationships among variables. In contrast, the CNN-based model ModernTCN~\cite{moderntcn} uses ConvFFN1 to independently learn new feature representations for each variable, while ConvFFN2 independently captures cross-variable dependencies for each feature. Additionally, Transformer-based models have demonstrated success in precise time series forecasting by capturing inter-variable interactions. Crossformer~\cite{CrossGNN} introduces a two-stage attention mechanism to capture both cross-time and cross-variable dependencies, significantly improving prediction accuracy. More recently, iTransformer embeds entire time series into variable tokens and employs explicit attention mechanisms to capture cross-variable dependencies, achieving state-of-the-art performance. However, the attention mechanism used in iTransformer~\cite{itransformer} fails to effectively capture complex and dynamic multivariate correlations and struggles to learn negative correlations between variables. To address these limitations, we propose a convolutional operation based on a timepoint-independent strategy that captures dynamic relationships, including both positive and negative correlations among variables.

\section{Proposed Method}\label{sec:proposed method}
\subsection{Problem Definition}
In multivariate time series forecasting tasks, given a historical input sequence 
\( \mathbf{X} = [\mathbf{x}^{(1)}, \mathbf{x}^{(2)}, \ldots, \mathbf{x}^{(L)}] \in \mathbb{R}^{L \times N} \),  the goal is to predict the future output sequence \( \mathbf{Y} = [\mathbf{x}^{(L+1)}, \mathbf{x}^{(L+2)}, \ldots, \mathbf{x}^{(L+T)}] \in \mathbb{R}^{T \times N} \), where \( N \) represents the number of variables, \( L \) and \( T \) respectively indicate the lengths of the input and output sequences. For convenience, $\mathbf{x}^{(i)} \in \mathbb{R}^{1 \times N}$ represents the values of $N$ variables at the $i$-th time point, where $x^{(i)}_j \in \mathbb{R}^{1 \times 1}$ denotes the $j$-th variable at the $i$-th time point.
\subsection{The General Structure}

The cross-variable correlation of multivariate time series demonstrates multifaceted (positive and negative correlation) and dynamic progression over time.
Considering this intrinsic nature of multivariate time series, we propose a TimeCNN model to refine cross-variable interaction on
time points to enhance prediction performance. 
Its key component is timepoint-independent, where each time point has an independent convolution kernel.
TimeCNN primarily consists of CrossCNN for capturing dynamic cross-variable dependencies on each time point, and Feed-Forward Networks (FFN) for learning generalizable representations to predict future time series. 

The overall architecture of TimeCNN is illustrated in Figure~\ref{fig:TimeCNN}. We first utilize CrossCNN to refine cross-variable interactions on time points.  Figure~\ref{fig:CrossCNN} gives the detailed structure of CrossCNN.
Specifically, for a sequence \( \mathbf{X} \in \mathbb{R}^{L \times N} \), we use an independent large convolution kernel for each time point, to learn the dependencies between \( N \) variables and obtain the output \( \mathbf{X}_{\text{crosscnn}} \in \mathbb{R}^{L \times N} \).
After cross-variable interaction, the acquired output is transposed, and we embed the entire time series of each variable into variable tokens individually. 
Subsequently, we utilize FFN to learn the time series representation for each variable.
We process the time series of each variable by sharing the same embedding and FFN.
Finally, the learned multivariate representations are transformed through Linear Projection to predict output \( \hat{\mathbf{Y}} \in \mathbb{R}^{T \times N} \).

The process for forecasting future sequences of \( \mathbf{Y} \), contingent
upon the historical series  \( \mathbf{X} \) can be succinctly delineated as follows:
\begin{align}
    &\mathbf{X}_{\text{crosscnn}} = \text{CrossCNN}(\mathbf{X}) , \\
    &\mathbf{X_{\text{trans}}} = \text{Transpose}(\mathbf{X}_{\text{crosscnn}}) ,  \\
    &\mathbf{X_{\text{emb}}} = \text{Embedding}(\mathbf{X_{\text{trans}}}), \\
    &\mathbf{X}_{\text{FFN}} = \text{FFN}(\mathbf{X_{\text{emb}}}), \\
    &\mathbf{\hat{Y}} = \text{Projection}(\mathbf{X}_{\text{FFN}}),
\end{align}
where \( \mathbf{X}_{\text{crosscnn}} \in \mathbb{R}^{L \times N} \), \( \mathbf{X_{\text{emb}}}, \mathbf{X}_{\text{FFN}} \in \mathbb{R}^{N \times D} \), and \( \mathbf{\hat{Y}} \in \mathbb{R}^{T \times N} \). \( N \) represents the number of variables, \( L \) and \( T \) represent the input and output step lengths, and \( D \) represents the feature dimensionality.


\subsection{Model Architecture}

\begin{figure*}[htbp]
    \centering
    \includegraphics[scale=0.50]{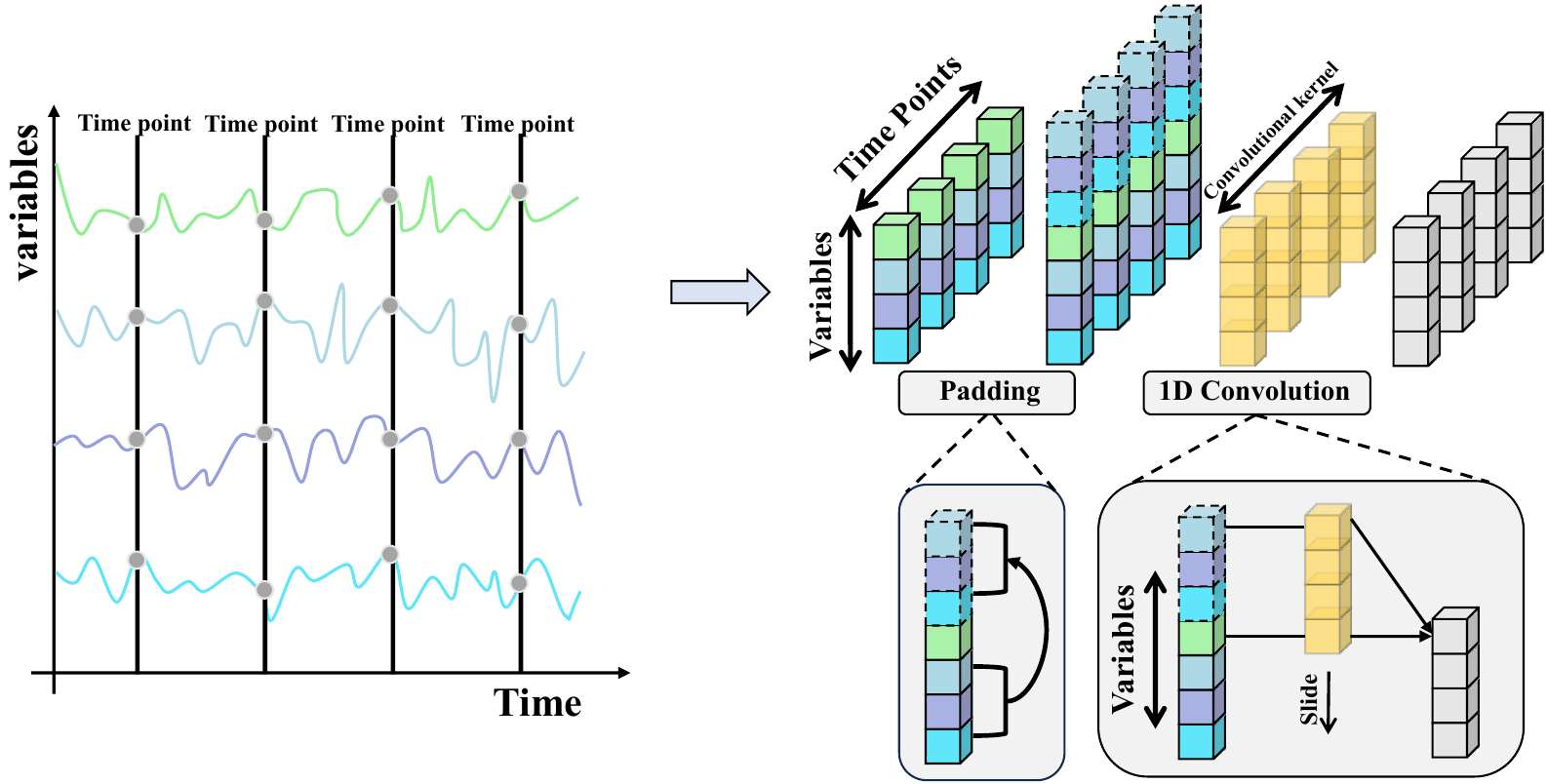}
    \caption{The CrossCNN architecture. Based on the timepoint-independent strategy, each time point has an independent convolutional kernel to refine cross-variable interaction.}
    \label{fig:CrossCNN}
\end{figure*}

\textbf{CrossCNN.} It adopts timepoint-independent, where each time point has an independent convolution kernel. The independent handling of time points enables CrossCNN to effectively capture dynamically evolving relationships among variables. Moreover, 
 this configuration ensures the modeling of all variables at each time point while minimizing the number of parameters to just $L \times N$ for all time points. 
The detailed structure of CrossCNN is shown in the Figure~\ref{fig:CrossCNN}. 
For the $i$-th time point input $\mathbf{x}^{(i)} \in \mathbb{R}^{1 \times N} = [x^{(i)}_1, x^{(i)}_2, \ldots, x^{(i)}_N]$, we pad it
 by taking the last $N-1$ variables and
 finally obtain   $\hat{\mathbf{x}}^{(i)} = [x^{(i)}_2, x^{(i)}_3, \ldots, x^{(i)}_N, x^{(i)}_1, x^{(i)}_2, \ldots, x^{(i)}_N]$,  as described in Figure~\ref{fig:CrossCNN}. 
After the padding operation, the sliding window mechanism of the convolution ensures that each convolutional operation covers all variables. 

Subsequently, the convolutional operation  
is performed to capture multivariate correlations at the $i$-th time point.
We set the convolutional kernel of size to the number of variables $N$ for enlarging the local receptive field.
The convolutional kernel parameters  are assumed  as $\mathbf{w}^{(i)} = [w^{(i)}_1, w^{(i)}_2, \ldots, w^{(i)}_N]$.
The $j$-th variable of the output, $c^{(i)}_j$, is obtained as follows:
\begin{equation}
c^{(i)}_j = \sum_{k=1}^{N} w^{(i)}_k \cdot \hat{x}^{(i)}_{(j+k) }, \quad j = 1, 2, \ldots, N, \quad i = 1, 2, \ldots, L.
\end{equation}
As demonstrated in Figure~\ref{fig:CrossCNN}, by sliding  convolutional operation we can obtain the output of the $i$-th time point $\mathbf{c}^{(i)} = [c^{(i)}_1, c^{(i)}_2, \ldots, c^{(i)}_N] \in \mathbb{R}^{1 \times N}$. We notice that the signs of the convolutional kernel parameters indicate the positive and negative correlations between variables at each point, which attention mechanisms fail to capture.
 
Finally, dropout is applied to the output via the convolutional operation to mitigate overfitting and promote model generalizability.
Furthermore, the production of CrossCNN
incorporates a skip connection to preserve the
integrity of the input data throughout the processing pipeline.

\textbf{Feed-Forward Networks (FFN).} 
In light of the advancements presented by DLinear~\cite{DLinear} and RLinear~\cite{ RLinear}, we employ  FFN to learn variable representations for time series forecasting. 
The acquired output $\mathbf{X}_{\text{crosscnn}}$ of CrossCNN
is transposed and the entire time series of each variable is subsequently embedded into individual variable tokens. Then, we obtain the input $\mathbf{X}_{\text{emb}}$  of FFN.
After cross-variable interaction,   the time series of each variable is processed by sharing the same embedding and FFN. 

As depicted in Figure~\ref{fig:TimeCNN}, to ensure consistent model training and improve the network's convergence properties, layer normalization is implemented at the input stage. 
Constituting a multi-layer perceptron (MLP) with a singular hidden layer, FFN employs a Gaussian Error Linear Unit (GELU) activation function to facilitate non-linear transformations. 
Furthermore, it incorporates a skip connection, characterized by its fully linear nature, to preserve the integrity of the input data throughout the processing pipeline. To mitigate overfitting and promote model generalizability, dropout is applied to the linear mapping between the hidden layer and the output. 
FFN consists of $M$ above MLP units and its formula is as follows:
\begin{align}
    \mathbf{X_{h}} &= \text{Dropout}(\text{Gelu}(\text{Dense}(\text{LayerNorm}( \mathbf{X_{m-1}})))),\\
    \mathbf{X_m} &= \text{Dropout}(\text{Dense}(\mathbf{X_h}))) +\mathbf{X_{m-1}}.
\end{align}
Here, $\mathbf{X_{m-1}}$ and $\mathbf{X_m} \in \mathbb{R}^{N \times D}$ are the input and output of the $m$-th layer, $\mathbf{X_h} \in \mathbb{R}^{N \times H}$, where $H$ represents the size of the hidden layer. We define $\mathbf{X_{0}}=\mathbf{X_{\text{emb}}}$  as the input of FFN.

\textbf{Projection Prediction.} 
A linear layer is employed to obtain the final prediction  \( \mathbf{\hat{Y}} = [\mathbf{\hat{x}}^{(L+1)}, \mathbf{\hat{x}}^{(L+2)}, \ldots, \mathbf{\hat{x}}^{(L+T)}] \in \mathbb{R}^{T \times N} \). 
We use Mean Squared Error (MSE) to measure the difference between the predicted values $\hat{\mathbf{Y}}$ and the ground truth $\mathbf{Y}$. MSE is calculated within $T$ time steps. The formula is as follows:
\begin{equation}
    \mathcal{L} = \frac{1}{T} \sum_{i=1}^{T} \left\| \hat{\mathbf{x}}^{(L+i)} - \mathbf{x}^{(L+i)} \right\|_{2}^{2}.
\end{equation}

\textbf{Instance Norm.}
As a commonly used preprocessing method for time series data, this approach is proposed to address the issue of distribution shift between the training and testing sets in some datasets~\cite{ulyanov2016instance}. Instance Norm subtracts the mean and divides it by the standard deviation for each single-variable sequence across all time steps. Subsequently, when obtaining the final predictions, denormalization will be applied.

\section{Experiments}
\subsection{Experiment Settings}\label{sec:experiment settings}
\subsubsection{Datasets}
We evaluate the performance of TimeCNN on 12 real-world datasets, including Weather, ECL, Traffic, ETT (ETTh1, ETTh2, ETTm1, ETTm2) used by AutoFormer~\cite{Autoformer}, Solar-Energy used by LST-Net~\cite{LSTNet} and PEMS datasets (PEMS03, PEMS04, PEMS07, PEMS08) adopted by SCINet~\cite{SCINet}. These are widely used multivariate time series datasets, and we handle the datasets the same way as iTransformer~\cite{itransformer}. The details of datasets are as follows:
\begin{itemize}
    \item \textbf{ETT (Electricity Transformer Temperature)}~\cite{Informer}\footnote{\href{https://github.com/zhouhaoyi/ETDataset}{https://github.com/zhouhaoyi/ETDataset}} includes four subsets. \textit{ETTh1 and ETTh2} collect hourly data on 7 different factors from two distinct electricity transformers from July 2016 to July 2018. \textit{ETTm1 and ETTm2} record the same factors at a higher resolution of every 15 minutes.
    \item \textbf{Traffic}~\cite{Autoformer}\footnote{\href{https://pems.dot.ca.gov/}{https://pems.dot.ca.gov/}} collects hourly data from the California Department of Transportation, describing road occupancy rates measured by 862 sensors on San Francisco Bay Area freeways.
    \item \textbf{ECL(Electricity)}~\cite{Autoformer}\footnote{\href{https://archive.ics.uci.edu/ml/datasets/ElectricityLoadDiagrams20112014}{https://archive.ics.uci.edu/ml/datasets/}} captures the hourly electricity consumption of 321 clients from 2012 to 2014.
    \item \textbf{Weather}~\cite{Autoformer}\footnote{\href{https://www.bgc-jena.mpg.de/wetter/}{https://www.bgc-jena.mpg.de/wetter/}} records 21 meteorological factors such as air temperature and humidity every 10 minutes throughout the year 2020.
    \item \textbf{Solar-Energy}~\cite{LSTNet}\footnote{\href{http://www.nrel.gov/grid/solar-power-data.html}{http://www.nrel.gov/grid/solar-power-data.html}} includes data from 137 PV plants in Alabama State, with solar power production sampled every 10 minutes during 2006.
    \item \textbf{PEMS}~\cite{SCINet}\footnote{\href{https://pems.dot.ca.gov/}{https://pems.dot.ca.gov/}} includes traffic network data from California, sampled every 5 minutes, focusing on four public subsets: \textit{PEMS03}, \textit{PEMS04}, \textit{PEMS07}, and \textit{PEMS08}.
\end{itemize}

The summary of the datasets is shown in Table~\ref{tab:dataset}.
\begin{table}[thbp]
  \caption{Summary of various datasets with details. Each dataset's description includes the number of variables, the proportions of data allocated to training, validation, and test sets, and the frequency of data collection.}\label{tab:dataset}
  \centering
  \begin{threeparttable}
  \renewcommand{\arraystretch}{1.2}
  \resizebox{\columnwidth}{!}{
  \begin{tabular}{l|c|c|c|c}
    \toprule
    Dataset & Variables & Proportion & Frequency & Domain \\
    \midrule
    ETTh1 & 7 & 6:2:2 & 1 hour & Electricity \\
    ETTh2 & 7 & 6:2:2 & 1 hour & Electricity \\
    ETTm1 & 7 & 6:2:2 & 15min & Electricity \\
    ETTm2 & 7 & 6:2:2 & 15min & Electricity \\
    Weather & 21 & 7:1:2 & 10min & Weather \\
    ECL & 321 & 7:1:2 & 1 hour & Electricity \\
    Traffic & 862 & 7:1:2 & 1 hour & Transportation \\
    Solar-Energy & 137 & 7:1:2 & 10min & Energy \\
    PEMS03 & 358 & 6:2:2 & 5min & Transportation \\
    PEMS04 & 307 & 6:2:2 & 5min & Transportation \\
    PEMS07 & 883 & 6:2:2 & 5min & Transportation \\
    PEMS08 & 170 & 6:2:2 & 5min & Transportation \\
    \bottomrule
  \end{tabular}
  }
  \end{threeparttable}
\end{table}

\subsubsection{Baselines} 
We select the state-of-the-art model iTransformer~\cite{itransformer} and 9 commonly recognized multivariate time series models as our baselines. Among them, Transformer-based models include iTransformer, PatchTST~\cite{PatchTST}, Crossformer~\cite{Crossformer}, FEDformer~\cite{fedformer}, and Autoformer~\cite{Autoformer}. Linear-based models include RLinear~\cite{RLinear}, DLinear~\cite{DLinear}, and TiDE~\cite{tide}. The CNN-based models include SCINet~\cite{SCINet} and TimesNet~\cite{Timesnet}. Each model adheres to a uniform experimental framework, characterized by a prediction interval, \(T\), which is defined within the set \(\{96, 192, 336, 720\}\).  
The lookback window is configured to 96.
We aggregate the results of baselines as reported by iTransformer~\cite{itransformer}, and the result of PatchTST and TimeCNN are from our experiments.

\subsubsection{Pseudocode of TimeCNN}
In the pseudocode presented in Algorithm~\ref{algo:TimeCNN}, we outline the architecture of TimeCNN, which processes an input time series through CrossCNN and a series of feed-forward neural network (FFN) layers to generate future predictions.

\begin{algorithm}
\caption{TimeCNN - Overall Architecture}\label{algo:TimeCNN}
\begin{algorithmic}[1]
\Require Input lookback time series $\mathbf{X}\in\mathbb{R}^{L\times N}$; input Length $L$; predicted length $T$; variates number $N$; token dimension $D$; FFN block number $M$.

\State $\mathbf{Padding}=\mathbf{X}[:, 1:]$
\Comment{select the first $N-1$ variables}

\State $\mathbf{Padding}= \texttt{Concatenate}(\mathbf{Padding}, \mathbf{X})$
\Comment{Concatenate along variable dimension, $\mathbf{Padding} \in \mathbb{R}^{L \times (2 \times N -1)}$}

\State $\mathbf{X}=\texttt{CrossCNN}(\mathbf{Padding}) + \mathbf{X}$
\Comment{CrossCNN Block, only one layer, and the size of each convolutional kernel is set to $N$.}

\State $\mathbf{X}=\mathbf{X}.\texttt{transpose}$
\Comment{$\mathbf{X} \in \mathbb{R}^{N \times L}$}

\State $\mathbf{H}^{0}=\texttt{MLP}(\mathbf{X})$
\Comment{Embed series into variable tokens}

\State $\mathbf{H}^{1} = \mathbf{H}^{0} + \texttt{FFN}(\mathbf{H}^{0})$
\Comment{Apply FFN block 1}

\State $\mathbf{H}^{2} = \mathbf{H}^{1} + \texttt{FFN}(\mathbf{H}^{1})$
\Comment{Apply FFN block 2}

\State \dots
\Comment{Repeat up to $M$ FFN blocks}

\State $\mathbf{H}^{M}=\texttt{LayerNorm}(\mathbf{H}^{M})$
\Comment{Normalize output}

\State $\mathbf{\hat{Y}}=\texttt{MLP}(\mathbf{H}^{M})$
\Comment{Project tokens to predicted series}

\State $\mathbf{\hat{Y}}=\mathbf{\hat{Y}}.\texttt{transpose}$
\Comment{$\mathbf{\hat{Y}} \in \mathbb{R}^{T \times N}$}

\Ensure $\mathbf{\hat{Y}}$
\Comment{Return the prediction result}
\end{algorithmic} 
\end{algorithm}

\subsection{Main Result}

\begin{table*}[htbp]
  \renewcommand{\arraystretch}{0.67}
  \centering
  \caption{Full results for the multivariable time series forecasting. We compare extensive existing models under four prediction lengths across various datasets. The lookback window is set to 96. The results of PatchTST and TimeCNN are from our experiments and the others are reported following iTransformer\cite{itransformer}.}
  \begin{threeparttable}
  \begin{small}
  \setlength{\tabcolsep}{2pt}
  \resizebox{\textwidth}{!}{
\begin{tabular}{c|c||cc|cc|cc|cc|cc|cc|cc|cc|cc|cc|cc}
    \toprule 
    \multicolumn{2}{c}{Models} &
    \multicolumn{2}{c}{\scalebox{0.8}{\textbf{TimeCNN}}} &
    \multicolumn{2}{c}{\scalebox{0.8}{iTransformer}} &
    \multicolumn{2}{c}{\scalebox{0.8}{RLinear}} &
    \multicolumn{2}{c}{\scalebox{0.8}{PatchTST}} &
    \multicolumn{2}{c}{\scalebox{0.8}{Crossformer}} &
    \multicolumn{2}{c}{\scalebox{0.8}{TiDE}} &
    \multicolumn{2}{c}{\scalebox{0.8}{TimesNet}} &
    \multicolumn{2}{c}{\scalebox{0.8}{DLinear}} &
    \multicolumn{2}{c}{\scalebox{0.8}{SCINet}} &
    \multicolumn{2}{c}{\scalebox{0.8}{FEDformer}} &
    \multicolumn{2}{c}{\scalebox{0.8}{Autoformer}} \\     
    \cmidrule(lr){3-4}
    \cmidrule(lr){5-6}
    \cmidrule(lr){7-8}
    \cmidrule(lr){9-10}
    \cmidrule(lr){11-12}
    \cmidrule(lr){13-14}
    \cmidrule(lr){15-16}
    \cmidrule(lr){17-18}
    \cmidrule(lr){19-20}
    \cmidrule(lr){21-22}
    \cmidrule(lr){23-24} 
    \multicolumn{2}{c}{Metric} & 
    \scalebox{0.78}{MSE} & \scalebox{0.78}{MAE} & 
    \scalebox{0.78}{MSE} & \scalebox{0.78}{MAE} & 
    \scalebox{0.78}{MSE} & \scalebox{0.78}{MAE} & 
    \scalebox{0.78}{MSE} & \scalebox{0.78}{MAE} & 
    \scalebox{0.78}{MSE} & \scalebox{0.78}{MAE} & 
    \scalebox{0.78}{MSE} & \scalebox{0.78}{MAE} & 
    \scalebox{0.78}{MSE} & \scalebox{0.78}{MAE} & 
    \scalebox{0.78}{MSE} & \scalebox{0.78}{MAE} & 
    \scalebox{0.78}{MSE} & \scalebox{0.78}{MAE} & 
    \scalebox{0.78}{MSE} & \scalebox{0.78}{MAE} & 
    \scalebox{0.78}{MSE} & \scalebox{0.78}{MAE} \\
\bottomrule
  \end{tabular}
  }
    \end{small}
  \end{threeparttable}
\label{tab:main_results}
\end{table*}

The experimental results are delineated in Table \ref{tab:main_results}, where lower MSE and MAE indicate better prediction accuracy. Our proposed TimeCNN outperforms all baselines on almost all datasets. Particularly, on datasets with a large number of variables, such as Traffic, Solar-Energy, and PEMS, TimeCNN exhibits significantly higher prediction accuracy compared to the cross-variable models iTransformer and Crossformer. This is mainly attributed to our strategy timepoint-independent, which enables TimeCNN to capture variable relationships at each time point using a large convolutional kernel.
Consequently, TimeCNN obtains the best performance by capturing the dynamically changing variable relationships over time. 
In contrast, iTransformer embeds entire time series into variable tokens, making it difficult to capture such dynamic relationships. Additionally, the attention mechanism of iTransformer and Crossformer fails to extract negative correlations between variables. On the other hand, compared to cross-time interaction models such as PatchTST, TimesNet, and DLinear, TimeCNN demonstrates significantly superior performance, alongside robust generalization capabilities across diverse datasets. Finally, our proposed TimeCNN can address real-world prediction challenges effectively.

\begin{table*}[htbp]
  \caption{Full results with a longer lookback window setting. The lookback window is set to the best prediction performance. The results of iTransformer~\cite{itransformer}, ModernTCN~\cite{moderntcn}, PatchTST~\cite{PatchTST}, RLinear~\cite{RLinear} and TimeCNN are from our experiments. And the other results are following ModernTCN.}
  \label{table:model_comparison}
  \centering
  \renewcommand{\arraystretch}{1.1}
  \scalebox{0.83}{
    \begin{tabular}{c@{\hspace{0.3em}}c@{\hspace{0.5em}}*{10}{|c@{\hspace{0.5em}}c}}
      \toprule
      \multirow{2}{*}{\textbf{Dataset}} & \multirow{2}{*}{\textbf{Horizon}} & 
      \multicolumn{2}{c}{\textbf{TimeCNN}} & 
      \multicolumn{2}{c}{iTransformer} &
      \multicolumn{2}{c}{ModernTCN} & 
      \multicolumn{2}{c}{PatchTST} & 
      \multicolumn{2}{c}{RLinear}   &
      \multicolumn{2}{c}{Crossformer} &
      \multicolumn{2}{c}{DLinear} &
      \multicolumn{2}{c}{FEDformer} &
      \multicolumn{2}{c}{TimesNet}&
      \multicolumn{2}{c}{SCINet}\\
      \cmidrule(lr){3-4} \cmidrule(lr){5-6} \cmidrule(lr){7-8} \cmidrule(lr){9-10} \cmidrule(lr){11-12}  \cmidrule(lr){13-14} \cmidrule(lr){15-16} \cmidrule(lr){17-18} \cmidrule(lr){19-20}  \cmidrule(lr){21-22}   
      & & MSE & MAE & MSE & MAE & MSE & MAE & MSE & MAE & MSE & MAE& MSE & MAE& MSE & MAE & MSE & MAE & MSE & MAE & MSE & MAE \\
      \midrule
      \multirow{4}{*}{Weather} 
      & 96  & \color{red}{\textbf{0.148}} & \color{red}{\textbf{0.196}} & 0.163 & 0.213 & 0.154 & 0.208 & \color{blue}{\underline{0.149}} & \color{blue}{\underline{0.198}} & 0.174 & 0.224 & 0.153 & 0.217 & 0.152 & 0.237 & 0.238 & 0.314 & 0.172 & 0.220 & 0.178 & 0.233 \\
      & 192 & \color{red}{\textbf{0.193}} & \color{red}{\textbf{0.239}} & 0.203 & 0.249 & 0.204 & 0.253 & \color{blue}{\underline{0.194}} & \color{blue}{\underline{0.241}} & 0.217 & 0.259 & 0.197 & 0.269 & 0.220 & 0.282 & 0.275 & 0.329 & 0.219 & 0.261 & 0.235 & 0.277\\
      & 336 & \color{red}{\textbf{0.243}} & \color{red}{\textbf{0.281}} & 0.255 & 0.290 & 0.251 & 0.287 & \color{blue}{\underline{0.245}} & \color{blue}{\underline{0.282}} & 0.264 & 0.293 & 0.252 & 0.311 & 0.265 & 0.319 & 0.339 & 0.337 & 0.280 & 0.306 & 0.337 & 0.345\\
      & 720 & \color{red}{\textbf{0.313}} & \color{red}{\textbf{0.334}} & 0.326 & 0.338 & \color{blue}{\underline{0.317}} & 0.335 & \color{blue}{\underline{0.314}} & \color{red}{\textbf{0.334}} & 0.331 & 0.339 & 0.318 & 0.363 & 0.323 & 0.362 & 0.389 & 0.409 & 0.365 & 0.359 & 0.396 & 0.413\\
      \midrule
      \multirow{4}{*}{Traffic}
      & 96  & \color{red}{\textbf{0.333}} & \color{red}{\textbf{0.244}} & \color{blue}{\underline{0.351}} & 0.257 & 0.368 &0.253 & 0.360 & \color{blue}{\underline{0.249}} & 0.410 & 0.279 & 0.512 & 0.290 & 0.410 & 0.282 & 0.576 & 0.359 & 0.593 & 0.321 & 0.613 & 0.395\\
      & 192 & \color{red}{\textbf{0.348}} & \color{red}{\textbf{0.252}} & \color{blue}{\underline{0.363}} & 0.264 & 0.379 &0.261 & 0.379 & \color{blue}{\underline{0.256}} & 0.423 & 0.284 & 0.523 & 0.297 & 0.423 & 0.287 & 0.610 & 0.380 & 0.617 & 0.336 & 0.559 & 0.363\\
      & 336 & \color{red}{\textbf{0.364}} & \color{red}{\textbf{0.260}} & \color{blue}{\underline{0.382}} & 0.274 & 0.397 &0.270 & 0.392 & \color{blue}{\underline{0.264}} & 0.434 & 0.290 & 0.530 & 0.300 & 0.436 & 0.296 & 0.608 & 0.375 & 0.629 & 0.336 & 0.555 & 0.358\\
      & 720 & \color{red}{\textbf{0.401}} & \color{red}{\textbf{0.278}} & \color{blue}{\underline{0.421}} & 0.293 & 0.440 &0.296 & 0.432 & \color{blue}{\underline{0.286}} & 0.464 & 0.307 & 0.573 & 0.313 & 0.466 & 0.315 & 0.621 & 0.375 & 0.640 & 0.350 & 0.620 & 0.394\\
      \midrule
      \multirow{4}{*}{ECL}
      & 96  & \color{red}{\textbf{0.125}} & \color{red}{\textbf{0.220}} & 0.130 & 0.226 & 0.131 & 0.226 & \color{blue}{\underline{0.129}} & \color{blue}{\underline{0.222}} & 0.141 & 0.236 & 0.187 & 0.283 & 0.153 & 0.237 & 0.186 & 0.302 & 0.168 & 0.272 & 0.171 & 0.256 \\
      & 192 & \color{blue}{\underline{0.146}} & \color{red}{\textbf{0.240}} & 0.153 & 0.250 & \color{red}{\textbf{0.145}} & \color{blue}{\underline{0.241}} & 0.147 & \color{red}{\textbf{0.240}} & 0.155 & 0.248 & 0.258 & 0.330 & 0.152 & 0.249 & 0.197 & 0.311 & 0.184 & 0.289 & 0.177 & 0.265 \\
      & 336 & \color{blue}{\underline{0.162}} & \color{red}{\textbf{0.257}} & 0.169 & 0.265 & \color{red}{\textbf{0.161}} & 0.260 & 0.163 & \color{blue}{\underline{0.259}} & 0.171 & 0.265 & 0.323 & 0.369 & 0.169 & 0.267 & 0.213 & 0.328 & 0.198 & 0.300 & 0.197 & 0.285 \\
      & 720 & \color{red}{\textbf{0.190}} & \color{red}{\textbf{0.287}} & \color{blue}{\underline{0.194}} & \color{blue}{\underline{0.288}} & \color{blue}{\underline{0.194}} & 0.290 & 0.197 & 0.290 & 0.210 & 0.297 & 0.404 & 0.423 & 0.233 & 0.344 & 0.233 & 0.344 & 0.220 & 0.320 & 0.234 & 0.318 \\
      \midrule
      \multirow{4}{*}{ETTh1}
      & 96  & \color{red}{\textbf{0.365}} & \color{blue}{\underline{0.394}} & 0.396 & 0.426 & 0.376 & 0.397 & 0.379 & 0.401 & \color{blue}{\underline{0.370}} & \color{red}{\textbf{0.392}} & 0.386 & 0.429 & 0.375 & 0.399 & 0.376 & 0.415 & 0.384 & 0.402 & 0.375 & 0.406   \\
      & 192 & \color{red}{\textbf{0.400}} & \color{blue}{\underline{0.415}} & 0.427 & 0.443 & 0.409 & 0.417 & 0.413 & 0.429 & 0.403 & \color{red}{\textbf{0.412}} & 0.419 & 0.444 & 0.405 & 0.416 & 0.423 & 0.446 & 0.557 & 0.436  & 0.416 & 0.402 \\
      & 336 & \color{red}{\textbf{0.422}} & \color{red}{\textbf{0.431}} & 0.457 & 0.467 & 0.437 & \color{blue}{\underline{0.434}} & 0.435 & 0.436 & 0.436 & \color{blue}{\underline{0.434}} & 0.440 & 0.461 & 0.439 & 0.443 & 0.444 & 0.462 & 0.491 & 0.469  & 0.504 & 0.495 \\
      & 720 & \color{red}{\textbf{0.442}} & \color{red}{\textbf{0.458}} & 0.503 & 0.491 & 0.457 & 0.466 & \color{blue}{\underline{0.446}} & 0.464 & 0.450 & \color{blue}{\underline{0.463}} & 0.519 & 0.527 & 0.472 & 0.490 & 0.469 & 0.492 & 0.521 & 0.500  & 0.544 & 0.527 \\
      \midrule 
        \multirow{4}{*}{ETTh2}
      & 96  & \color{red}{\textbf{0.267}} & \color{red}{\textbf{0.332}} & 0.299 & 0.359 & 0.275 & 0.341 & 0.274 & \color{blue}{\underline{0.337}} & \color{blue}{\underline{0.273}} & 0.338 & 0.628 & 0.563 & 0.289 & 0.353 & 0.332 & 0.374 & 0.340 & 0.374 & 0.295 & 0.361  \\
      & 192 & \color{red}{\textbf{0.328}} & \color{red}{\textbf{0.374}} & 0.377 & 0.406 & 0.341 & 0.388 & \color{blue}{\underline{0.338}} & \color{blue}{\underline{0.376}} & 0.342 & 0.386 & 0.703 & 0.624 & 0.383 & 0.418 & 0.407 & 0.446 & 0.402 & 0.414 & 0.349 & 0.383  \\
      & 336 & \color{red}{\textbf{0.353}} & \color{blue}{\underline{0.398}} & 0.415 & 0.438 & 0.367 & 0.412 & \color{blue}{\underline{0.363}} & \color{red}{\textbf{0.397}} & 0.365 & 0.409 & 0.827 & 0.675 & 0.448 & 0.465 & 0.400 & 0.447 & 0.452 & 0.452 & 0.365 & 0.409  \\
      & 720 & \color{red}{\textbf{0.391}} & \color{red}{\textbf{0.429}} & 0.416 & 0.443 & 0.407 & 0.440 & \color{blue}{\underline{0.393}} & \color{blue}{\underline{0.430}} & 0.434 & 0.455 & 1.181 & 0.840 & 0.605 & 0.551 & 0.412 & 0.469 & 0.462 & 0.468 & 0.475 & 0.488  \\
      \midrule
       \multirow{4}{*}{ETTm1}
      & 96  & \color{red}{\textbf{0.284}} & \color{red}{\textbf{0.342}} & 0.302 & 0.357 & 0.298 & 0.349 & \color{blue}{\underline{0.293}} & 0.346 & 0.305 & 0.346 & 0.316 & 0.373 & 0.299 & \color{blue}{\underline{0.343}} & 0.326 & 0.390 & 0.338 & 0.375 & 0.325 & 0.372     \\
      & 192 & \color{red}{\textbf{0.325}} & \color{red}{\textbf{0.363}} & 0.345 & 0.383 & 0.335 & 0.372 & \color{blue}{\underline{0.333}} & 0.370 & 0.339 & 0.366 & 0.377 & 0.411 & 0.335 & \color{blue}{\underline{0.365}} & 0.365 & 0.415 & 0.371 & 0.387 & 0.354 & 0.386   \\
      & 336 & \color{red}{\textbf{0.357}} & \color{blue}{\underline{0.385}} & 0.379 & 0.403 & 0.373 & 0.396 & \color{blue}{\underline{0.369}} & 0.397 & 0.370 & \color{red}{\textbf{0.383}} & 0.431 & 0.442 & \color{blue}{\underline{0.369}} & 0.386 & 0.392 & 0.425 & 0.410 & 0.411 & 0.394 & 0.415    \\
      & 720 & \color{red}{\textbf{0.410}} & \color{blue}{\underline{0.417}} & 0.439 & 0.437 & 0.420 & 0.419 & \color{blue}{\underline{0.416}} & 0.420 & 0.427 & \color{red}{\textbf{0.415}} & 0.600 & 0.547 & 0.425 & 0.421 & 0.446 & 0.458 & 0.478 & 0.450  & 0.476 & 0.469   \\
      \midrule
         \multirow{4}{*}{ETTm2}
      & 96  & \color{red}{\textbf{0.161}} & \color{red}{\textbf{0.252}} & 0.175 & 0.266 & 0.171 & 0.257 & 0.166 & 0.256 & \color{blue}{\underline{0.164}} & \color{blue}{\underline{0.253}} & 0.421 & 0.461 & 0.167 & 0.260 & 0.180 & 0.271 & 0.187 & 0.267 & 0.186 & 0.281 \\
      & 192 & \color{red}{\textbf{0.215}} & \color{red}{\textbf{0.289}} & 0.238 & 0.312 & 0.224 & 0.295 & 0.223 & 0.296 & \color{blue}{\underline{0.218}} & \color{blue}{\underline{0.290}} & 0.503 & 0.519 & 0.224 & 0.303 & 0.252 & 0.318 & 0.249 & 0.309 & 0.277 & 0.356 \\
      & 336 & \color{red}{\textbf{0.268}} & \color{red}{\textbf{0.325}} & 0.289 & 0.341 & \color{blue}{\underline{0.272}} & 0.327 & 0.274 & 0.329 & 0.273 & \color{blue}{\underline{0.326}} & 0.611 & 0.580 & 0.281 & 0.342 & 0.324 & 0.364 & 0.321 & 0.351 & 0.311 & 0.369 \\
      & 720 & \color{red}{\textbf{0.344}} & \color{red}{\textbf{0.379}} & 0.377 & 0.398 & \color{blue}{\underline{0.349}} & 0.388 & 0.362 & \color{blue}{\underline{0.385}} & 0.369 & 0.386 & 0.996 & 0.750 & 0.397 & 0.421 & 0.410 & 0.420 & 0.497 & 0.403 & 0.403 & 0.412 \\
      \midrule
    \multicolumn{2}{c}{\textbf{1$^{st}$ Count}}  & \color{red}{\textbf{26}} &\color{red}{\textbf{23}} &0 &0 &2 &0 &0 &3 &0 &0 &4 &0 &0 &0 &0 &0 &0 &0 &0 &0 \\
      \bottomrule
    \end{tabular}
}
\end{table*}

\begin{table*}[htbp]
\setlength{\abovecaptionskip}{5pt} 
\setlength{\belowcaptionskip}{0pt} 
\caption{Efficiency comparison among models on the Weather, Electricity, PEMS03 and PEMS07 datasets, where MACs represent the number of multiply-accumulate operations, Params denotes the model's parameter count, and Time indicates the inference time.}
\label{tab:efficiency_analysis}
\centering
\resizebox{\textwidth}{!}{
\renewcommand{\arraystretch}{1.5}
\begin{tabular}{c|ccc|ccc|ccc|ccc}
\hline
& \multicolumn{3}{c|}{Weather}  
& \multicolumn{3}{c|}{Electricity} 
& \multicolumn{3}{c|}{PEMS03} 
& \multicolumn{3}{c}{PEMS07} \\ \hline
Methods         & MACS      & Params    & Time
                & MACS      & Params    & Time
                & MACS      & Params    & Time 
                & MACS      & Params    & Time \\ \hline
TimeCNN        & \textbf{\color{red}26.88M}    & 1.279M    & \textbf{\color{red}0.25ms}  
                & \textbf{\color{red}1.10G}     & 3.409M   & \textbf{\color{red}0.32ms}       
                & \textbf{\color{red}0.80G}     & 2.217M    & \textbf{\color{red}0.38ms} 
                & \textbf{\color{red}2.01G}     & 2.267M    & \textbf{\color{red}0.38ms} \\
iTransformer    & 123.85M   & 4.957M     & 0.67ms 
                & 1.61G   & 4.957M     & 0.67ms 
                & 2.31G     & 5.552M     & 0.84ms     
                & 5.89G    & 6.386M     & 2.53ms     \\
PatchTST        & 111.22M   & \textbf{\color{red}915.95K}     & 0.73ms    
                & 1.70G     & \textbf{\color{red}915.95K}     & 1.05ms    
                & 2.30G     & \textbf{\color{red}601.26K}     & 0.95ms
                & 5.68G     & \textbf{\color{red}601.26K}     & 1.74ms \\
ModernTCN       & 57.29M   & 2.450M     & 0.45ms    
                & 3.24G     & 128.671M     & 3.78ms    
                & 3.78G     & 156.559M     & 3.89ms
                & 20.72G     &860.748M     & 29.51ms \\ \hline
\end{tabular}}

\end{table*}

In addition, we extended our experiments by utilizing a longer lookback window setting to further evaluate the performance of our model. For comparison, we selected several state-of-the-art models as baselines, including iTransformer~\cite{itransformer}, ModernTCN~\cite{moderntcn}, PatchTST~\cite{PatchTST}, RLinear~\cite{RLinear}, Crossformer~\cite{Crossformer}, DLinear~\cite{DLinear}, FEDformer~\cite{fedformer}, and SCINet~\cite{SCINet}. During the evaluation process, we corrected a bug related to the setting 'dropout\_last = True' in ModernTCN, PatchTST, and RLinear, which could have otherwise led to erroneous prediction performance, especially when utilizing a larger batch size as a hyperparameter. Our lookback window setting is aligned with ModernTCN, following its configuration to optimize prediction performance.
The results for Crossformer, DLinear, FEDformer, and SCINet were obtained from the ModernTCN paper, while the results for the other models were generated through our own experimental runs. As presented in Table~\ref{table:model_comparison}, the experimental results demonstrate that, even with an extended lookback window, our proposed TimeCNN consistently outperforms these baseline models. We attribute this superior performance to the strong scalability of our CrossCNN block, which is highly effective in extracting cross-variable dependencies at each time point over a longer lookback window, significantly enhancing prediction accuracy.

\subsection{Efficiency Analysis}\label{sec:efficiency analysis}
\subsubsection{With Fixed Setting}
We conducted experiments to evaluate the efficiency metrics of TimeCNN, including Multiply-Accumulate Operations (MACs)~\cite{MACs}, the number of parameters (Params), and inference time (Time). We evaluated these metrics on the Weather and ECL datasets using prediction lengths of 96, 192, 336, and 720 time steps, and on the PEMS03 and PEMS07 datasets using prediction lengths of 12, 24, 48, and 96 time steps, with a lookback window of 96 time steps and a batch size of 1 across all experiments. Results are averaged from all prediction lengths. All experiments are conducted on a GeForce RTX 4090 24GB. Notably, during inference time testing, we conducted 300 iterations of GPU preheating and average over 10,000 trials to ensure precision. The methodology of calculating inference time is detailed at the site\footnote{https://deci.ai/blog/measure-inference-time-deep-neural-networks/}. And we selected iTransformer~\cite{itransformer}, PatchTST~\cite{PatchTST} and ModernTCN~\cite{moderntcn} as baseline models for comparison.

The results are shown in the Table~\ref{tab:efficiency_analysis}. TimeCNN provides a better balance of efficiency and performance. Across the four datasets used in our experiments, TimeCNN achieves a 60.46\% reduction in MACs and a 57.50\% reduction in the number of parameters compared to iTransformer, while achieves a speed that is 3 to 4 times faster than iTransformer. This demonstrates that TimeCNN surpasses iTransformer in both prediction accuracy and computational efficiency. 

Although PatchTST employs variable-independent strategies to ensure that each variable uses the same Transformer-Encoder blocks, thereby reducing the parameter count, this does not result in a reduction in MACs and significantly increases inference time. In contrast, TimeCNN exhibits fewer MACs (approximately 60.41\% reduction) and faster inference times (about 3 to 4 times faster). Moreover, TimeCNN achieves significantly higher prediction accuracy compared to PatchTST, especially on datasets with numerous variables.

On the other hand, compared to the CNN-based model ModernTCN, TimeCNN achieves a 72.16\% reduction in MACs and an 85.87\% reduction in the number of parameters. This is primarily due to ModernTCN’s small patch sizes and higher hidden layer dimensions, which lead to an explosion in computational cost, particularly on datasets with numerous variables, and result in much longer inference times. Furthermore, across the four datasets, TimeCNN is, on average, nearly 25 times faster than ModernTCN. Overall, our experiments highlight TimeCNN as an effective and efficient model for time series forecasting tasks.

\begin{figure*}[htbp]
    \centering
    \hspace*{-0.8cm} 
    \includegraphics[scale=0.5]{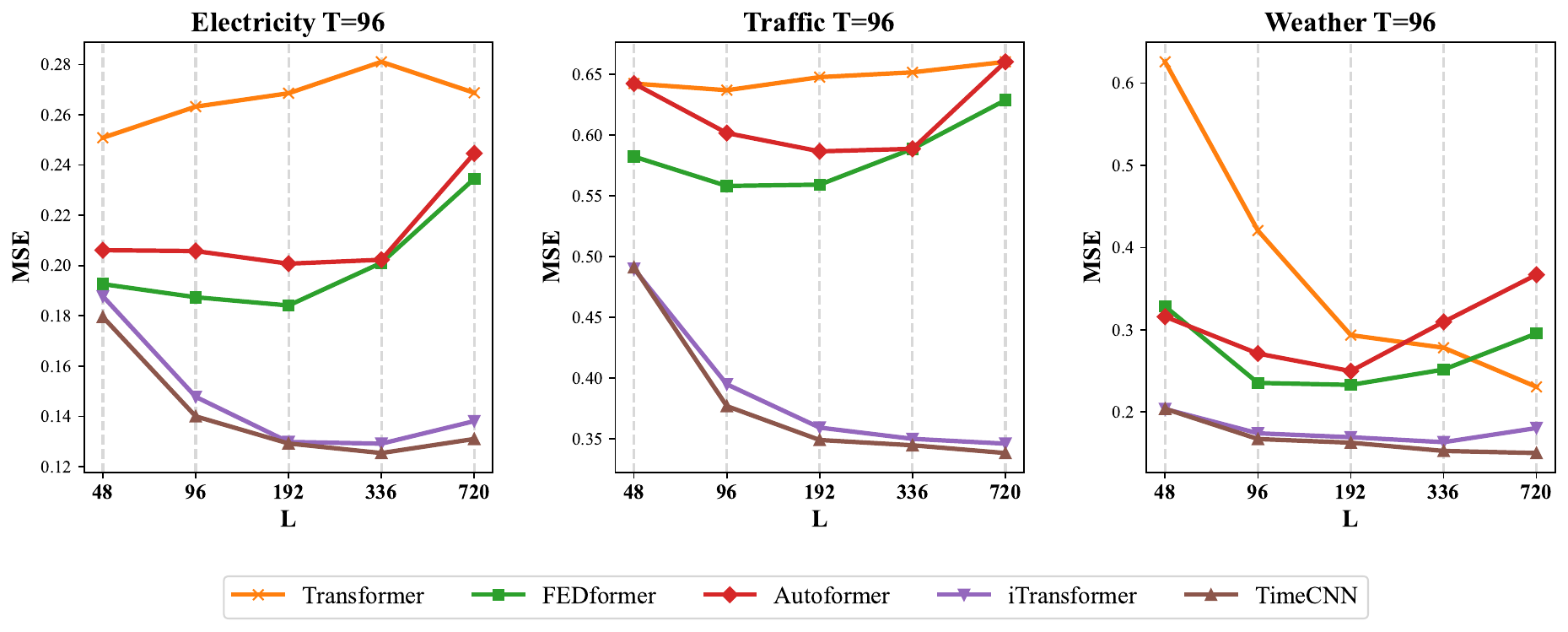}
    \caption{Sensitivity Analysis of Lookback Window Length $L$. On the ECL, Traffic, and Weather datasets, we set $L \in \{48, 96, 192, 336, 720\}$, and the prediction length $T = 96$. The accuracy of TimeCNN increases with the increasing lookback window, surpassing other models.}
    \label{fig:increasing_length}
\end{figure*}

\subsubsection{With Various Setting}
\begin{figure}[htbp]
    \centering
   \includegraphics[width=\columnwidth]{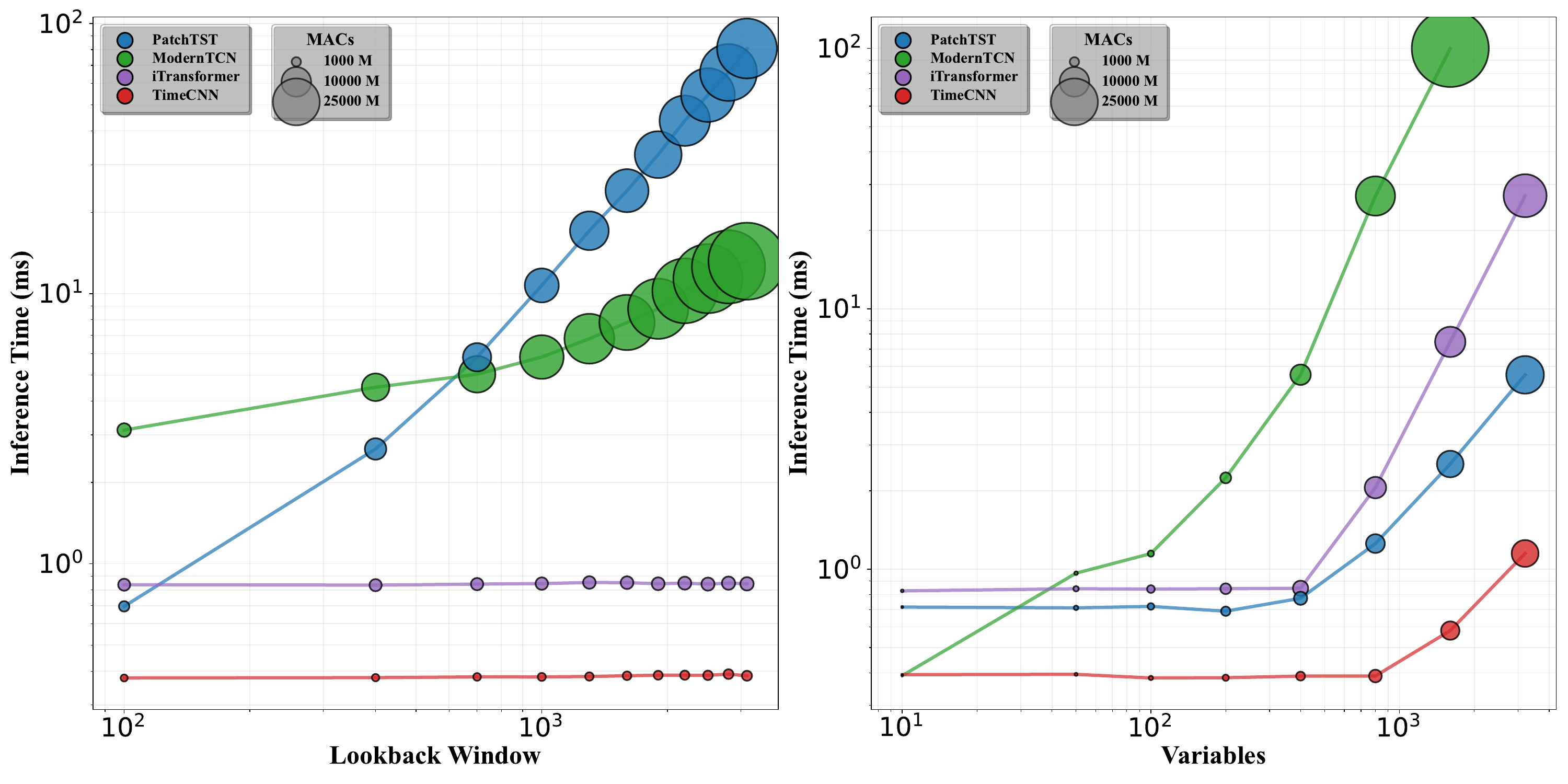}
    \caption{The efficiency with various lookback window or variable numbers.}
    \label{fig:efficiency}
\end{figure}

We configured various lookback windows and adjusted the number of variables to assess the efficiency of our TimeCNN model in comparison to other baselines. The results are presented in Figure~\ref{fig:efficiency}. The left panel illustrates the changes in Multiply-Accumulate Operations (MACs) and Inference Time under different lookback window configurations, while the right panel shows the corresponding metrics for different numbers of variables. Across all scenarios, our TimeCNN consistently achieves lower computational overhead and faster inference times. This improvement arises primarily due to the computational complexity of PatchTST growing quadratically with the length of the lookback window (with a fixed patch size), while iTransformer's Computational complexity scales quadratically with the number of variables. Furthermore, ModernTCN encounters computational and memory inefficiencies due to higher embedding dimensions and a larger number of patches, resulting in memory exhaustion when the number of variables exceeds 1000, even with a batch size of 1 on a GeForce RTX 4090 24GB GPU. In contrast, TimeCNN, by utilizing convolutional operations to extract variable information independently on each time point, significantly reduces computational costs through parallelism. This also suggests that TimeCNN exhibits superior robustness to varying multivariate input sequences, achieving both higher efficiency and improved performance.

\subsection{Model Analysis }

\subsubsection{Increasing Lookback Window}
Previous work~\cite{DLinear} has indicated that most Transformer-based models such as Informer~\cite{Informer}, Autoformer~\cite{Autoformer}, and FEDformer~\cite{fedformer} struggle to attain more precise future predictions from increasing lookback windows.
However, longer lookback windows imply larger receptive fields. We also conducted experiments to test the performance of our proposed TimeCNN across different historical windows. The experimental results are illustrated in the Figure~\ref{fig:increasing_length}. On datasets with a larger number of variables such as Weather, ECL, and Traffic, TimeCNN's predictive performance surpasses these models across almost all benchmarks. Furthermore, TimeCNN can capture dynamic variable relationships with longer lookback windows to enhance predictive capabilities.

\begin{table*}[htbp]
\caption{Performance enhancement achieved by incorporating our CrossCNN block. We report the average performance and the relative MSE and MAE reduction (Promotion).}
\label{table:2}
\resizebox{\textwidth}{!}{
\renewcommand{\arraystretch}{1.25}  
\begin{tabular}{c|c|cc|cc|cc|cc|cc}
\hline
\multicolumn{2}{c|}{Models} & \multicolumn{2}{c|}{ECL} & \multicolumn{2}{c|}{Traffic} & \multicolumn{2}{c|}{PEMS03} & \multicolumn{2}{c|}{PEMS04} & \multicolumn{2}{c}{PEMS07} \\
\hline
\multicolumn{2}{c|}{Metric} & MSE & MAE & MSE & MAE & MSE & MAE & MSE & MAE & MSE & MAE \\
\hline
\multirow{3}{*}{RMLP} & Original & 0.190 & 0.277 & 0.515 & 0.331 & 0.155 & 0.263 & 0.173 & 0.281 & 0.171 & 0.267 \\
 & \textbf{+CrossCNN} & \textbf{\color{red}0.183} & \textbf{\color{red}0.274} & \textbf{\color{red}0.476} & \textbf{\color{red}0.321} & \textbf{\color{red}0.116} & \textbf{\color{red}0.231} & \textbf{\color{red}0.112} & \textbf{\color{red}0.227} & \textbf{\color{red}0.105} & \textbf{\color{red}0.217} \\
 & Promotion & \textbf{3.68\%} & \textbf{1.08\%} & \textbf{7.57\%} & \textbf{3.02\%} & \textbf{25.16\%} & \textbf{12.17\%} & \textbf{35.37\%} & \textbf{19.22\%} & \textbf{38.60\%} & \textbf{18.73\%} \\
\hline
\multirow{3}{*}{PatchTST} & Original & 0.190 & 0.275 & 0.467 & 0.293 & 0.136 & 0.241 & 0.150 & 0.258 & 0.150 & 0.247 \\
 & \textbf{+CrossCNN} & \textbf{\color{red}0.180} & \textbf{\color{red}0.272} & \textbf{\color{red}0.434} & \textbf{\color{red}0.286} & \textbf{\color{red}0.109} & \textbf{\color{red}0.221} & \textbf{\color{red}0.111} & \textbf{\color{red}0.225} & \textbf{\color{red}0.100} & \textbf{\color{red}0.211} \\
 & Promotion & \textbf{5.26\%} & \textbf{1.09\%} & \textbf{7.07\%} & \textbf{2.39\%} & \textbf{19.85\%} & \textbf{8.30\%} & \textbf{26.00\%} & \textbf{12.79\%} & \textbf{33.33\%} & \textbf{14.57\%} \\
\hline
\multirow{3}{*}{iTransformer} & Original & 0.178 & 0.270 & 0.428 & 0.282 & 0.113 & 0.222 & 0.104 & 0.208 & 0.101 & 0.204 \\
 & \textbf{+CrossCNN} & \textbf{\color{red}0.171} & \textbf{\color{red}0.267} & \textbf{\color{red}0.417} & \textbf{\color{red}0.279} & \textbf{\color{red}0.107} & \textbf{\color{red}0.214} & \textbf{\color{red}0.093} & \textbf{\color{red}0.198} & \textbf{\color{red}0.086} & \textbf{\color{red}0.186} \\
 & Promotion & \textbf{3.93\%} & \textbf{1.11\%} & \textbf{2.57\%} & \textbf{1.06\%} & \textbf{5.31\%} & \textbf{3.60\%} & \textbf{10.58\%} & \textbf{4.81\%} & \textbf{14.85\%} & \textbf{8.82\%} \\
\hline
\end{tabular}
}
\label{tab:with CrossCNN}
\end{table*}

\begin{table*}[htbp]
\caption{The ablation study of CrossCNN and FFN. We remove the CrossCNN module or replace it with a linear layer to test the performance of CrossCNN and FFN. The experimental results are the average of four prediction lengths.}
\label{tab:random_seed}
\centering
\renewcommand{\arraystretch}{1.5} 
\resizebox{\textwidth}{!}{
\begin{tabular}{c|cc|cc|cc|cc|cc}
\hline
Dataset & \multicolumn{2}{c|}{ECL} & \multicolumn{2}{c|}{Traffic} & \multicolumn{2}{c|}{Weather}& \multicolumn{2}{c|}{ETTm1} & \multicolumn{2}{c}{PEMS03}\\
Horizon & MSE      & MAE             & MSE      & MAE             & MSE      & MAE         & MSE      & MAE      & MSE      & MAE            \\ \hline
96 or 12      & 0.140$\pm$0.000 & 0.236$\pm$0.000 & 0.377$\pm$0.000 & 0.257$\pm$0.000 & 0.167$\pm$0.001 & 0.210$\pm$0.001 & 0.315$\pm$0.002 & 0.351$\pm$0.001 & 0.062$\pm$0.000 & 0.165$\pm$0.000 \\
192 or 24     & 0.156$\pm$0.000 & 0.250$\pm$0.000 & 0.398$\pm$0.001 & 0.267$\pm$0.000 & 0.215$\pm$0.000 & 0.254$\pm$0.001 & 0.360$\pm$0.001 & 0.377$\pm$0.000 & 0.080$\pm$0.000 & 0.187$\pm$0.000 \\
336 or 48     & 0.172$\pm$0.001 & 0.267$\pm$0.001 & 0.416$\pm$0.001 & 0.274$\pm$0.000 & 0.272$\pm$0.001 & 0.296$\pm$0.001 & 0.396$\pm$0.001 & 0.401$\pm$0.000 & 0.115$\pm$0.001 & 0.225$\pm$0.001 \\
720 or 96     & 0.212$\pm$0.003 & 0.304$\pm$0.002 & 0.451$\pm$0.001 & 0.293$\pm$0.000 & 0.355$\pm$0.001 & 0.351$\pm$0.000 & 0.460$\pm$0.001 & 0.439$\pm$0.000 & 0.163$\pm$0.001 & 0.269$\pm$0.001 \\ \hline

\end{tabular}}
\end{table*}

\subsubsection{Module Ablation}
We conducted ablation experiments to assess the contributions of the CrossCNN and FFN components within TimeCNN. In these experiments, we either replaced CrossCNN with alternative CNN variants or removed it entirely to evaluate its effectiveness. Specifically, the 'CrossLinear' replaces CrossCNN with a linear layer for variable mixing, following the approach used in TSMixer~\cite{TSMixer}, where fully connected layers capture dependencies across both variables and time. The 'OneCNN' employs a single convolutional kernel applied uniformly across all time points, using the same padding and operations as in TimeCNN. Additionally, the '2DCNN\_3' and '2DCNN\_7' substitute CrossCNN with (3, 3) and (7, 7) convolutional kernels, respectively, designed to capture temporal relationships and variable dependencies. Furthermore, we introduced a 'w/o CrossCNN' model, where the CrossCNN block is removed entirely. In all cases, the number of layers is kept constant, with the CrossCNN block replaced. We also added PatchTST~\cite{PatchTST} as a baseline in our ablation study. The lookback window is configured to 96, with prediction lengths set to 96, 192, 336, and 720, respectively.

\begin{figure}[htbp]
    \centering
   \includegraphics[width=\columnwidth]{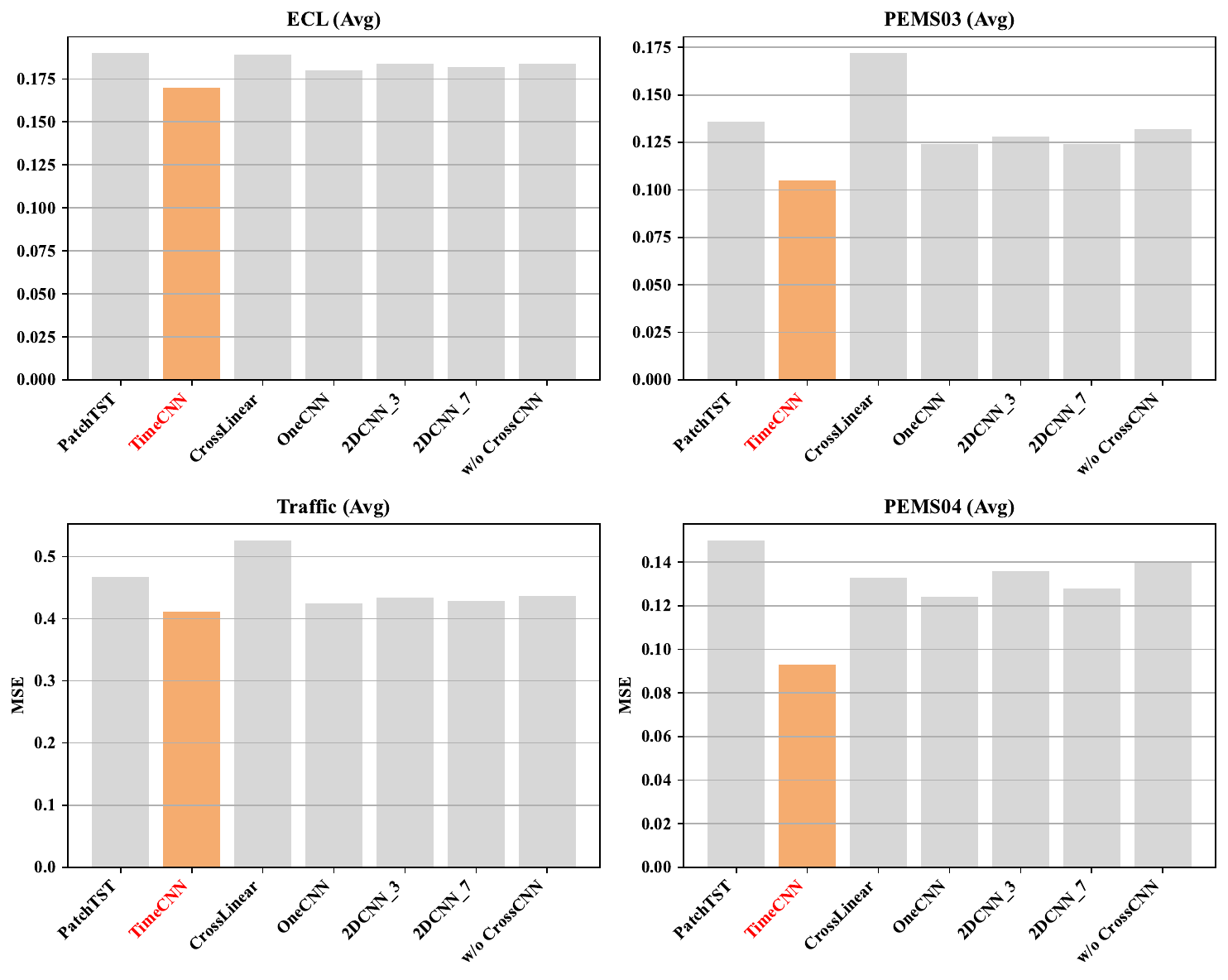}
    \caption{Ablation study of the components of TimeCNN}
    \label{fig:ablation_plots}
\end{figure}

The experimental results, as presented in Table~\ref{fig:ablation_plots}, demonstrate that TimeCNN consistently outperforms other models across all datasets. Specifically, TimeCNN surpasses CrossLinear in terms of MSE, showing an average reduction of 25.1\%, and MAE, with a decrease of 16.5\%. This improvement is primarily attributed to CrossLinear’s suboptimal performance on datasets with a larger number of variables, where the quadratic increase in parameters as variables grow leads to overfitting. In contrast, our proposed CrossCNN employs a shared convolutional kernel across all variables at each time step, resulting in a linear increase in parameters, which significantly reduces the risk of overfitting and enhances predictive accuracy. Furthermore, OneCNN applies a single convolution across all time points, overlooking the intrinsic dynamic relationships between variables over time, similar to the limitation observed in iTransformer. As a result, its performance is inferior to that of TimeCNN. Although 2D convolutional kernels can capture variable relationships to a certain extent, their limited receptive fields hinder their ability to fully model inter-variable dependencies, which further explains their lower performance compared to TimeCNN. Notably, even without the CrossCNN block, the model achieves superior predictive accuracy compared to PatchTST across all datasets, indicating that the FFN component is effective in modeling cross-time interactions and learning variable representations.

\subsubsection{Performance promotion}
The CrossCNN block is a modular component designed to capture dynamic relationships between variables and can be seamlessly integrated into any time series forecasting models. To assess its effectiveness as a supplementary module, we incorporated CrossCNN into recent baseline models, including iTransformer~\cite{itransformer}, PatchTST~\cite{PatchTST}, and RMLP~\cite{RLinear}. Specifically, we insert the CrossCNN block immediately after instance normalization to capture cross-variable relationships over time, while keeping all other components unchanged. The lookback window was set to 96, and the prediction lengths were set to 96, 192, 336 and 720. The experimental results, as shown in Table~\ref{tab:with CrossCNN}, indicate performance improvements across all datasets when CrossCNN is added to the baselines, particularly for datasets with a large number of variables. RMLP and PatchTST, which are variable-independent models, do not inherently capture inter-variable relationships. However, the inclusion of CrossCNN significantly enhances their performance by capturing these dynamic relationships. Similarly, in the case of iTransformer, using CrossCNN to capture inter-variable dependencies leads to performance gains. We attribute this modest improvement to the fact that the attention mechanism in iTransformer, to some extent, mitigates the dynamic relationships captured by CrossCNN.

\subsection{Robustness study}
\subsubsection{Random Seeds} 
We conducted a series of experiments with different random seeds to assess the robustness of the TimeCNN model. In both the main experiments and the other ablation study, we initially used random seed 2023. To further evaluate robustness, we performed multiple trials with random seeds 2021, 2022, 2023, 2024, and 2025, using the ECL, Traffic, Weather, ETTm1 and PEMS03 datasets. For all experiments, the input lengths were set to 96 and the prediction lengths were set to the same as the main experiment. We calculate the standard deviation of TimeCNN performance under five runs. The results are shown in Figure~\ref{tab:random_seed}, the results indicate that the performance of TimeCNN remains stable across different random seeds, demonstrating its robustness in handling variations in initialization.

\begin{figure}[htbp]
    \centering
    \includegraphics[scale=0.35]{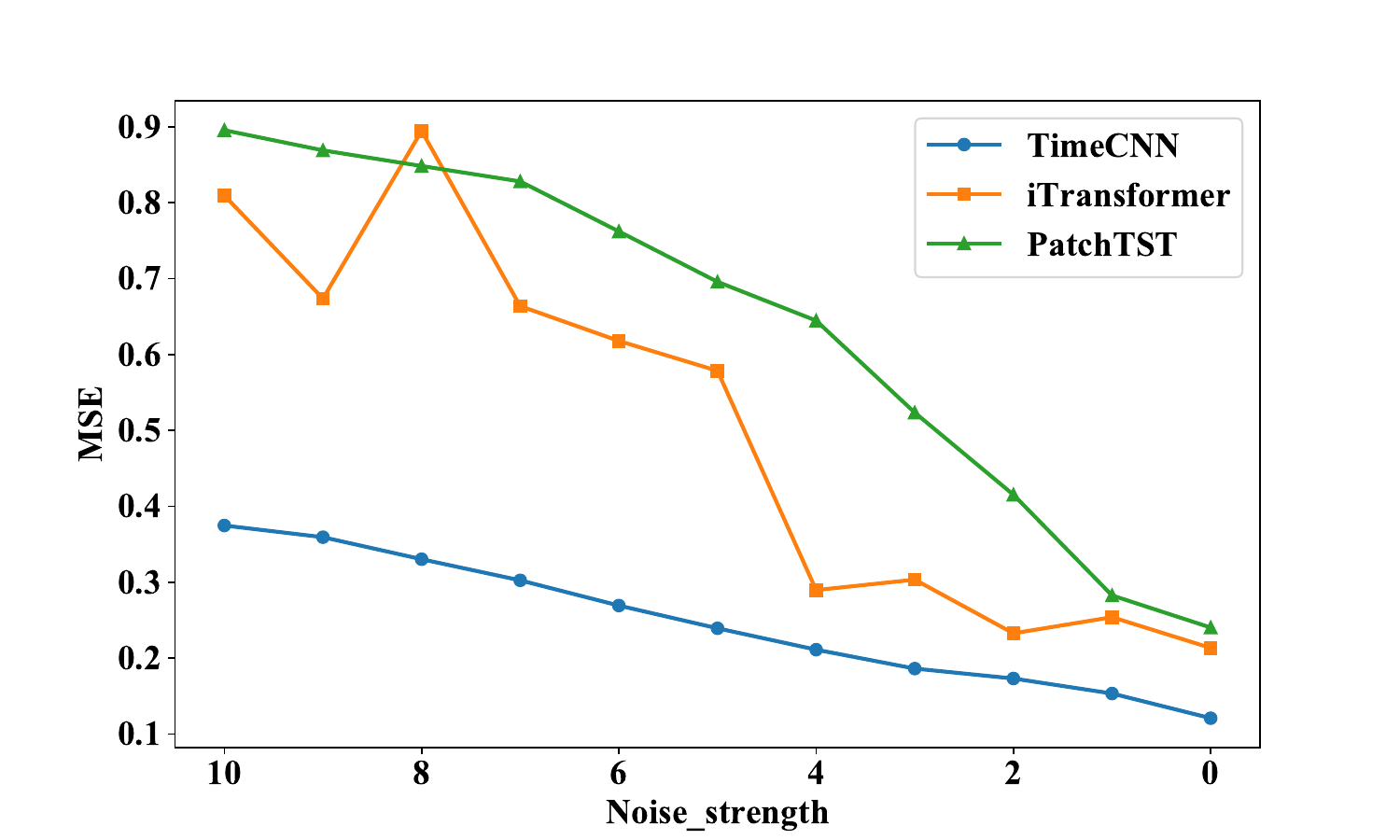}
    \caption{The robustness study of noise}
    \label{fig:noise}
\end{figure}

\subsubsection{Variable Noise}
We conducted experiments on the PEMS04 dataset to assess the robustness of TimeCNN in handling noisy input data. Specifically, we selected a single variable from the input data and introduced Gaussian noise with a mean of 0 and a standard deviation of $Noise\_strength$, applied at various intensities. After training under normal conditions, we calculated the mean squared error (MSE) during testing, focusing only on the predictions of the selected variable as the experimental result. The lookback window and prediction length were both set to 96. We used iTransformer~\cite{itransformer} and PatchTST~\cite{PatchTST} as baseline models for comparison. The experimental results are presented in Figure~\ref{fig:noise}. The results indicate that as the noise intensity increases, PatchTST’s MSE rises sharply. This is because the strategy of variable-independent processing makes it difficult for PatchTST to capture the inherent patterns in noisy time series. Although iTransformer outperforms PatchTST, its performance becomes unstable as noise intensifies, likely due to the susceptibility of attention scores to noise interference. In contrast, our proposed TimeCNN demonstrates better and more stable performance as noise intensity increases. This improvement can be attributed to the dynamic interactions between variables, which significantly mitigate the impact of noise.

\subsection{Instance Visualization}
\begin{figure}[htbp]
    \centering
    \includegraphics[scale=0.5]{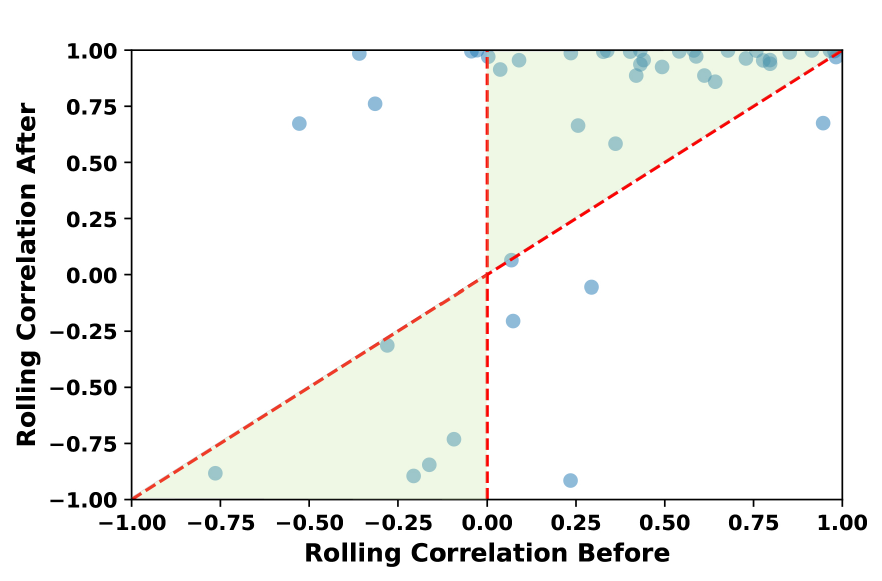}
    \caption{The instance visualization from PEMS08}
    \label{fig:valid}
\end{figure}

Our CrossCNN is designed to capture both positive and negative relationships among variables over time. Specifically, our CrossCNN can enhance the existing correlations between variables. For instance, variables that initially exhibit positive correlation demonstrate stronger positive correlation after information extraction through CrossCNN. To demonstrate this effect, before and after applying CrossCNN we calculate the rolling correlation coefficients between two variables over time. We select the latter half of the input time series (with a length of 48) and set a rolling window of 4. The resulting scatter plot of correlation coefficients is presented in Figure~\ref{fig:valid}.
The scatter plot reveals that the majority of points fall within the shaded regions, indicating that originally positively correlated variables show strengthened positive correlations, while originally negatively correlated variables exhibit enhanced negative correlations. This observation validates CrossCNN's capability to capture dynamically changing variable relationships over time.

\section{Conclusion And Future Work}\label{sec:conclusion}
Our research demonstrates that TimeCNN effectively addresses the limitations of current Transformer-based models in capturing the intricate and dynamic interactions among variable correlations in multivariate time series forecasting. The innovative architecture of TimeCNN, featuring timepoint-independent convolutional kernels, allows the model to independently learn cross-variable dependencies at each time point, adn capture both positive and negative correlations among variables. A comprehensive evaluation across 12 real-world datasets confirms that TimeCNN outperforms existing state-of-the-art models in predictive performance while achieving substantial improvements in computational efficiency. These results highlight the potential of TimeCNN as a robust and efficient advancement in the field of time series forecasting.

\ifCLASSOPTIONcaptionsoff
  \newpage
\fi



%

\printbibliography

@article{STGCN,  
 title={Spatio-Temporal Graph Convolutional Networks: A Deep Learning Framework for Traffic Forecasting}, 
 journal={IJCAI}, 
 author={Yu, Bing and Yin, Haoteng and Zhu, Zhanxing}, 
 year={2018},
 }

@article{MTGNN ,  
 title={Connecting the Dots: Multivariate Time Series Forecasting with Graph Neural Networks}, 
 journal={Proceedings of the 26th ACM SIGKDD International Conference on Knowledge Discovery Data Mining}, 
 author={Wu, Zonghan and Pan, Shirui and Long, Guodong and Jiang, Jing and Chang, Xiaojun and Zhang, Chengqi}, 
 year={2020}
 }

@book{weather_apply,
  title={Empirical Orthogonal Functions and Statistical Weather Prediction},
  author={Lorenz, E.N. and Statistical Forecasting Project (Massachusetts Institute of Technology)},
  year={1956},
  journal={Massachusetts Institute of Technology, Department of Meteorology}
}

@article{He_Zhang_Bai_Yi_Niu_2022, 
    title={CATN: Cross Attentive Tree-Aware Network for Multivariate Time Series Forecasting}, 
    journal={Proceedings of the AAAI Conference on Artificial Intelligence}, 
    author={He, Hui and Zhang, Qi and Bai, Simeng and Yi, Kun and Niu, Zhendong}, 
    year={2022} }

@article{RePEc:eee:rensus:v:74:y:2017:i:c:p:902-924,
  author={Deb, Chirag and Zhang, Fan and Yang, Junjing and Lee, Siew Eang and Shah, Kwok Wei},
  title={{A review on time series forecasting techniques for building energy consumption}},
  journal={Renewable and Sustainable Energy Reviews},
  year=2017,
}

@article{finance_apply,
   title={A time series analysis-based stock price prediction using machine learning and deep learning models},
   journal={International Journal of Business Forecasting and Marketing Intelligence},
   author={Mehtab, Sidra and Sen, Jaydip},
   year={2020},
 }

@article{Timesnet,
  title={TimesNet: Temporal 2D-Variation Modeling for General Time Series Analysis},
  author={Wu, Haixu and Hu, Tengge and Liu, Yong and Zhou, Hang and Wang, Jianmin and Long, Mingsheng},
  journal={ICLR},
  year={2023}
}

@article{PatchTST,
  title={A Time Series is Worth 64 Words: Long-term Forecasting with Transformers},
  author={Nie, Yuqi and Nguyen, Nam H and Sinthong, Phanwadee and Kalagnanam, Jayant},
  journal={ICLR},
  year={2023}
}

@article{itransformer,
  title={iTransformer: Inverted Transformers Are Effective for Time Series Forecasting},
  author={Liu, Yong and Hu, Tengge and Zhang, Haoran and Wu, Haixu and Wang, Shiyu and Ma, Lintao and Long, Mingsheng},
  journal={ICLR},
  year={2024}
}

@article{Crossformer,
  title={Crossformer: Transformer utilizing cross-dimension dependency for multivariate time series forecasting},
  author={Zhang, Yunhao and Yan, Junchi},
  journal={ICLR},
  year={2023}
}

@article{SCINet,
  title={SCINet: time series modeling and forecasting with sample convolution and interaction},
  author={Liu, Minhao and Zeng, Ailing and Chen, Muxi and Xu, Zhijian and Lai, Qiuxia and Ma, Lingna and Xu, Qiang},
  journal={NeurIPS},
  year={2022}
}

@article{Autoformer,
  title={Autoformer: Decomposition Transformers with {Auto-Correlation} for Long-Term Series Forecasting},
  author={Haixu Wu and Jiehui Xu and Jianmin Wang and Mingsheng Long},
  journal={NeurIPS},
  year={2021}
}

@article{Informer,
  title={Informer: Beyond efficient transformer for long sequence time-series forecasting},
  author={Li, Jianxin and Hui, Xiong and Zhang, Wancai},
  journal={arXiv: 2012.07436},
  year={2021}
}

@article{fedformer,
  title={{FEDformer}: Frequency enhanced decomposed transformer for long-term series forecasting},
  author={Zhou, Tian and Ma, Ziqing and Wen, Qingsong and Wang, Xue and Sun, Liang and Jin, Rong},
  journal={ICML},
  year={2022}
}

@article{DLinear,
  title={Are Transformers Effective for Time Series Forecasting?},
  author={Ailing Zeng and Muxi Chen and Lei Zhang and Qiang Xu},
  journal={AAAI},
  year={2023}
}

@article{Transformer,
 author = {Vaswani, Ashish and Shazeer, Noam and Parmar, Niki and Uszkoreit, Jakob and Jones, Llion and Gomez, Aidan N and Kaiser, Lukasz and Polosukhin, Illia},
 journal = {NeurIPS},
 title = {Attention is All you Need},
 year = {2017}
}

@article{LayerNorm,
 author = {Ba, Jimmy Lei  and  Kiros, Jamie Ryan  and  Hinton, Geoffrey E.},
 journal = {https://arxiv.org/pdf/1607.06450.pdf},
 title = {Layer Normalization},
 year = {2016}
}

@article{LSTNet,
  title={Modeling long-and short-term temporal patterns with deep neural networks},
  author={Lai, Guokun and Chang, Wei-Cheng and Yang, Yiming and Liu, Hanxiao},
  journal={SIGIR},
  year={2018}
}

@article{TSMixer,
  title={TSMixer: Lightweight MLP-Mixer Model for Multivariate Time Series Forecasting},
  author={Vijay Ekambaram and Arindam Jati and Nam Nguyen and Phanwadee Sinthong and Jayant Kalagnanam},
  journal={KDD},
  year={2023}
}

@article{RLinear,
  title={Revisiting Long-term Time Series Forecasting: An Investigation on Linear Mapping},
  author={Li, Zhe and Qi, Shiyi and Li, Yiduo and Xu, Zenglin},
  journal={arXiv preprint arXiv:2305.10721},
  year={2023}
}

@article{CrossGNN,
  title={CrossGNN: Confronting Noisy Multivariate Time Series Via Cross Interaction Refinement},
  author={Huang, Qihe and Shen, Lei and Zhang, Ruixin and Ding, Shouhong and Wang, Binwu and Zhou, Zhengyang and Wang, Yang},
  journal={NeurIPS},
  year={2023},
}

@article{ulyanov2016instance,
  title={Instance normalization: The missing ingredient for fast stylization},
  author={Ulyanov, Dmitry and Vedaldi, Andrea and Lempitsky, Victor},
  journal={arXiv preprint arXiv:1607.08022},
  year={2016}
}

@article{BaiTCN2018,
  author={Shaojie Bai and J. Zico Kolter and Vladlen Koltun},
  title={An Empirical Evaluation of Generic Convolutional and Recurrent Networks for Sequence Modeling},
  journal={arXiv:1803.01271},
  year={2018}
}

@article{tide,
  author={Abhimanyu Das and Weihao Kong and Andrew Leach and Shaan Mathur and Rajat Sen and Rose Yu},
  title={Long-term Forecasting with TiDE: Time-series Dense Encoder}, 
  journal={arXiv:2304.08424},
  year={2024}
}

@article{MACs,
  author={Xuelin Cao and Bo Yang and Chongwen Huang and George C. Alexandropoulos and Chau Yuen and Zhu Han and H. Vincent Poor and Lajos Hanzo},
  title={Massive Access of Static and Mobile Users via Reconfigurable Intelligent Surfaces: Protocol Design and Performance Analysis},
  journal={{IEEE} J. Sel. Areas Commun.},
  year={2022}
}

@article{moderntcn,
  title={ModernTCN: A Modern Pure Convolution Structure for General Time Series Analysis},
  author={Donghao Luo and Xue Wang},
  journal={ICLR},
  year={2024},
}

@article{Weather_TKDE,
  author={Yongshun Gong and Tiantian He and Meng Chen and Bin Wang and Liqiang Nie and Yilong Yin},
  journal={IEEE Transactions on Knowledge and Data Engineering}, 
  title={Spatio-Temporal Enhanced Contrastive and Contextual Learning for Weather Forecasting}, 
  year={2024},
}

@article{FInance_TKDE,
  author={Yu Zhao and Huaming Du and Ying Liu and Shaopeng Wei and Xingyan Chen and Fuzhen Zhuang and Qing Li and Gang Kou},
  journal={IEEE Transactions on Knowledge and Data Engineering}, 
  title={Stock Movement Prediction Based on Bi-Typed Hybrid-Relational Market Knowledge Graph via Dual Attention Networks}, 
  year={2023},
}

@article{DNN1_TKDE,
  author={Zhangjing Yang and Weiwu Yan and Xiaolin Huang and Lin Mei},
  journal={IEEE Transactions on Knowledge and Data Engineering}, 
  title={Adaptive Temporal-Frequency Network for Time-Series Forecasting}, 
  year={2022},
}

@article{DNN2_TKDE,
  author={Ming Jin and Yu Zheng and Yuan-Fang Li and Siheng Chen and Bin Yang and Shirui Pan},
  journal={IEEE Transactions on Knowledge and Data Engineering}, 
  title={Multivariate Time Series Forecasting With Dynamic Graph Neural ODEs}, 
  year={2023},
}

@ARTICLE{wen,
  author={Wen, Liangjian and Hu, Quan and Guo, Cong and Hu, Ao and Zhang, Mingyi},
  journal={IEEE Signal Processing Letters}, 
  title={Cross-Scale Attention for Long-term Time Series Forecasting}, 
  year={2024},
}

\end{document}